\documentclass[12pt]{article}

\newtheorem{theorem}{Theorem}[section]
\newtheorem{definition}{Definition}[section]
\newtheorem{lemma}{Lemma}[section]

\newtheorem{remark}{Remark}[section]

\usepackage{bm}
\usepackage[numbers]{natbib}
\usepackage{graphicx}
\usepackage{amssymb,amsmath}
\oddsidemargin
-0.2in \topmargin -.60in
\begin{document}

\title{Task-group Relatedness and Generalization Bounds for Regularized Multi-task Learning}

\author{Chao~Zhang\thanks{C.~Zhang is with the School of Mathematical Sciences, Dalian University of Technology,
Dalian, Liaoning, 116024, P.R. China. (e-mail: chao.zhang@dlut.edu.cn).},  Dacheng~Tao\thanks{D.~Tao is with the Centre for Quantum Computation
\& Intelligent Systems, FEIT, University of Technology, Sydney, NSW 2007, Australia. (e-mail: dacheng.tao@gmail.com).}, Tao~Hu\thanks{T.~Hu is with the School of Mathematical Sciences, Capital Normal University, Beijing, 100048 , P.R. China. (e-mail: hutaomath@foxmail.com).}, 
Xiang~Li\thanks{X.~Li is with the School of Mathematical Sciences, Dalian University of Technology,
Dalian, Liaoning, 116024, P.R. China. (e-mail: lixiangalixiang@gmail.com).}  % <-this % stops a space
%\thanks{This work is supported by XXXXXXXX. }% <-this % stops a space
}

\maketitle

\begin{abstract}
In this paper, we study the generalization performance of regularized multi-task learning (RMTL) in a vector-valued framework, where MTL is considered as a learning process for vector-valued functions. We are mainly concerned with two theoretical questions: 1) under what conditions does RMTL perform better with a smaller task sample size than STL? 2) under what conditions is RMTL generalizable and can guarantee the consistency of each task during simultaneous learning?
 In particular, we investigate two types of task-group relatedness: the observed discrepancy-dependence measure (ODDM) and the empirical discrepancy-dependence measure (EDDM), both of which detect the dependence between two groups of
multiple related tasks (MRTs). We then introduce the Cartesian product-based uniform entropy number (CPUEN) to measure the complexities of vector-valued function classes. By applying the specific deviation and the symmetrization inequalities to the vector-valued framework, we obtain the generalization bound for RMTL, which is the upper bound of the joint probability of the event that there is at least one task with a large empirical discrepancy between the expected and empirical risks. Finally, we present a sufficient condition to guarantee the consistency of each task in the simultaneous learning process, and we discuss how task relatedness affects the generalization performance of RMTL. Our theoretical findings answer the aforementioned two questions.
\end{abstract}


\section{Introduction}

There is plenty of empirical evidence to suggest that task-relatedness information
improves multi-task learning (MTL) over single-task learning (STL) in multiple related task (MRT) scenarios. Therefore, capturing relatedness information is important for both theoretical and practical investigations of MTL.

Several learning methods have been proposed to address this problem. \citet{evgeniou2005learning} introduced regularized MTL to link the simultaneous learning process of MRT scenarios to STL problems, in which the regularization terms encode the relatedness between MRTs. However, regularization term design relies on a priori knowledge of tasks. Other methods that model task relatedness let the different tasks share common structures, {\it e.g.,} backpropagation networks \citep{caruana1997multitask} and the structure learning formulation \citep{ando2005framework}. \citet{argyriou2007multi} presented a method to learn a low-dimensional representation shared across MRTs, while
\citet{Zhang10convex} applied covariance to model three types of relatedness between two tasks: the positive correlation, the negative correlation, and unrelatedness.
From the theoretical standpoint, the notion ``$\mathcal{F}$-related" has been proposed to study the generalizability of multi-task classification, where if two tasks $\mathcal{Z}^{[1]},\mathcal{Z}^{[2]}$ are $\mathcal{F}$-related for a given function class $\mathcal{F}$, there exists a function $f\in\mathcal{F}$ such that $P^{[1]}=f(P^{[2]})$ or $P^{[2]}=f(P^{[1]})$ \citep{ben2008notion,ben2003exploiting}. The interested reader is also referred to other theoretical investigations of MTL \citep{maurer2006bounds,maurer2013sparse} and learning theory \citep{mendelson2008lower,cesa2008improved,agarwal2013generalization,zhou2003capacity,jin2013improved,hussain2011design}.

\subsection{Overview of Main Results}

As discussed by \citet{micchelli2004kernels,micchelli2005learning}, MTL can be studied from the viewpoint of vector-valued function learning. Inspired by \citep{micchelli2004kernels,micchelli2005learning}, we explore the vector-valued framework to study the generalization and consistency properties of regularized MTL (RMTL) and analyze the relationship between the properties of RMTL and task-group relatedness. In particular, we address the following theoretical questions:
\begin{itemize}
\item Under what conditions does RMTL perform better with a smaller task sample size than STL?

\item Under what conditions is RMTL generalizable and can guarantee the consistency of each task during simultaneous learning?
\end{itemize}
In order to answer these questions, we also need to consider: 1) measures of task-group relatedness; 2) the joint probability of MRTs; 3) measures of
vector-valued function classes; and 4) the specific deviation and symmetrization inequalities for the vector-valued framework.

Here, we introduce two types of task-group relatedness: the observed discrepancy-dependence measure (ODDM) and the empirical discrepancy-dependence measure (EDDM) (see Section \ref{sec:measure}).\footnote{In this paper, the observed discrepancy is defined as the discrepancy between an observation and its expectation, and the empirical discrepancy is defined as the discrepancy between the expectation ({\it i.e.}, expected risk) and its empirical estimate ({\it i.e.}, empirical risk).} ODDM measures the statistical dependence between events that some tasks have large observed discrepancies and the others have small observed discrepancies. EDDM measures the statistical dependence between events that some tasks have large empirical discrepancies and the others have small empirical discrepancies. In contrast to ODDM, EDDM reflects the asymptotic behavior of the relatedness between two task groups when the sample size goes to {\it infinity}.\footnote{For convenience, we assume that all tasks have the same sample size in this paper.} We show that ODDM (or EDDM) can exist in three states: negative, positive, and {\it zero}, which respectively model three types of relatedness between two task groups: the synergy effect, the negative synergy effect, and unrelatedness.

Since MTL refers to a process in which MRTs are simultaneously processed, we consider the task joint probability, defined in \eqref{eq:joint.prob}, instead of the task summation probability as in \citep{maurer2006bounds,maurer2013sparse,ando2005framework,evgeniou2004regularized}. In task joint probability, the generalization bound for MTL is deemed to be the upper bound of the joint probability that there is at least one task with a large empirical discrepancy in MTL. This bound can also be used to describe the consistency of each task in the MTL learning process. In order to obtain the bound, we present the specific deviation inequalities and the symmetrization inequalities for the vector-valued framework and, meanwhile, introduce the Cartesian product-based uniform entropy number (CPUEN), which is induced from the uniform entropy numbers (UENs) of MRTs.

Based on the resulting generalization bounds, the theoretical properties of RMTL are analyzed and we show that:
\begin{itemize}
%
%\item Theoretically, the group-type relatedness is enough to guarantee the generalization performance of MTL.\vspace{-1mm}
%
\item the validity of RMTL will theoretically be guaranteed if most of the relatedness between two task groups show a synergy effect. If almost any pair of task groups are predominantly mutual, RMTL performs well with less samples than STL, and the required sample size of each task in RMTL will not increase dramatically, regardless of the (large) number of MRTs  (see Remarks \ref{rem:ODDM1}\&\ref{rem:ODDM2}).

\item there will be a tighter generalization bound for RMTL if the values of EDDMs are negative, {\it i.e.}, if most of the relatedness between two task groups show a synergy effect. Moreover, we present a sufficient condition to guarantee the consistency of each task in RMTL.

\end{itemize}
Furthermore, we obtain the following theoretical findings:
\begin{itemize}
  \item The aforementioned sufficient condition can be used to examine whether the given tasks, function classes, and regularization terms are suitable for MTL.
  \item The existence of a negative correlation between two tasks is necessary for MTL, which is in accordance with the argument by \citet{Zhang10convex}.
  \item The generalization bound of RMTL.
 \item	The relationship between the task relatedness and the generalization performance of RMTL.
 \item	The sufficient condition to guarantee the consistency of each task in RMTL.
\item The proposed vector-valued framework can be used to study the theoretical properties of vector-valued function learning \citep{micchelli2005learning}

\end{itemize}

%and we refer to \citet{micchelli2005learning} for the details on learning vector-valued functions.
%
%It is noteworthy that this scheme has been widely applied to to many existing models of multi-task learning, for example, the backpropagation networks \citep{caruana1997multitask}, the kernel models \citep{micchelli2004kernels,evgeniou2004regularized}, sparse-coding models \citep{maurer2013sparse}, multilinear models \citep{romera2013multilinear} and regularized linear models \citep{zhou2013modeling,zhou2011multi,zhou2011clustered}.

%
\subsection{Organization of the Paper}

The rest of this paper is organized as follows. In Section \ref{sec:setup}, the main research addressed in this paper, including the task-joint probability and generalization bounds for RMTL, is formalized. In Section \ref{sec:measure}, two quantities for measuring task-group relatedness are presented and CPUEN is introduced in Section \ref{sec:UEN} to measure the complexity of the vector-valued function classes. The main results are presented in Section \ref{sec:main}, along with a method to examine the validity of MTL. In Section \ref{sup:covariance}, we address the generalization performance results using the covariance information of MRTs and the last section concludes the paper.
In Appendix, we first present the deviation inequalities and the symmetrization inequalities for the vector-valued framework (Parts \ref{sup:deviation} \& \ref{sup:symmetrization}). Finally, the proofs of the main results are given in Part \ref{sup:proofs}.

%-----------------------------------------------------

\section{Problem Setup}\label{sec:setup}

We first formalize the main research addressed in this paper, including the task-joint probability and generalization bounds for RMTL.
%
%For convenience, a list of notations is given in Tab. \ref{tab:notation}
%
%\begin{table}[htbp]
%\centering
%\caption{A list of notations}
%\begin{tabular}{|c|c|}
%\hline
%Math Notation & Meaning \\
%\hline
% $\mathcal{X}^{[m]}$& input space of the $m$-th task\\
%\hline
%$\mathcal{Y}^{[m]}$& output space of the $m$-th task\\
%\hline
%$\mathcal{Z}^{[m]}$& Cartesian product of $\mathcal{X}^{[m]}\times\mathcal{Y}^{[m]}$\\
%\hline
%$\mathcal{G}^{[m]}$& function class of the $m$-th task\\
%\hline
%\end{tabular}\label{tab:notation}
%\end{table}

\subsection{Regularized Multi-task Learning}

Given a space $\mathcal{X}\subset\mathbb{R}^I$, let $\mathcal{X}^{[m]}$ be the input space of the $m$-th task with the probability distribution $\mathcal{D}^{[m]}$ on $\mathcal{X}$ and $\mathcal{Y}^{[m]}\in\mathbb{R}^J$ be the corresponding output space ($1\leq m\leq M$). Let $g^{[m]}_{*}:\mathcal{X}^{[m]}\rightarrow \mathcal{Y}^{[m]}$ be the corresponding labeling function. Also, denote the $m$-th task as $\mathcal{Z}^{[m]}:=\mathcal{X}^{[m]}\times \mathcal{Y}^{[m]}\subset\mathbb{R}^K$ with $K=I+J$.

In MTL, let $\mathcal{G}^{[1]},\cdots,\mathcal{G}^{[M]}\subset\mathcal{Y}^{\mathcal{X}}$ be $M$ function classes corresponding to the learning tasks $\mathcal{Z}^{[1]},\cdots,\mathcal{Z}^{[M]}$, respectively. MTL is expected to simultaneously find $M$ functions $\widetilde{g}^{[1]}, \cdots,\widetilde{g}^{[M]}$ from $\mathcal{G}^{[1]},\cdots,\mathcal{G}^{[M]}$
such that each $\widetilde{g}^{[m]}$ can minimize the expected risk of the corresponding task $\mathcal{Z}^{[m]}$ over $\mathcal{G}^{[m]}$:
\begin{equation}\label{eq:exrisk}
    \mathrm{E}^{[m]}(\ell^{[m]} \circ g^{[m]})= \int\ell^{[m]}(g^{[m]}({\bf x}^{[m]}),{\bf y}^{[m]})d\mathrm{P}^{[m]}({\bf z}^{[m]}),\;\;1\leq m\leq M,
\end{equation}
where $\ell^{[m]}$ and $P^{[m]}({\bf z}^{[m]})$ are the loss function and the probability distribution of the task $\mathcal{Z}^{[m]}$, respectively, with ${\bf z}^{[m]}:=({\bf x}^{[m]},{\bf y}^{[m]})^T$.

Since the task distributions $P^{[1]},\cdots,P^{[M]}$ are usually unknown, the target functions $\widetilde{g}^{[1]}, \cdots,\widetilde{g}^{[M]}$ cannot be directly obtained by minimizing the expected risks \eqref{eq:exrisk} of MRTs. Instead, the empirical risk minimization (ERM) principle can be used to handle this issue. For each task $\mathcal{Z}^{[m]}$, let ${\bf Z}_{N}^{[m]}:=\{{\bf z}^{[m]}_n\}_{n=1}^{N}$ be a set of $N$ i.i.d. samples drawn from $\mathcal{Z}^{[m]}$ with ${\bf z}_n^{[m]}:=({\bf x}_n^{[m]},{\bf y}_n^{[m]})^T$.  The following is the objective function of RMTL:
\begin{equation*}%\label{eq:optimization2}
 \sum_{m=1}^M \mathrm{E}^{[m]}_N(\ell^{[m]}\circ g^{[m]}) +r \mathrm{R}(g^{[1]},\cdots,g^{[M]}),
\end{equation*}
where
\begin{equation}\label{eq:em.risk}
  \mathrm{E}^{[m]}_N(\ell^{[m]}\circ g^{[m]}):=\frac{1}{N}\sum_{n=1}^N\ell^{[m]}(g({\bf x}^{[m]}_n),{\bf
    y}^{[m]}_n),
\end{equation}
is the empirical risk of the task $\mathcal{Z}^{[m]}$, $\mathrm{R}(g^{[1]},\cdots,g^{[M]})$ is the regularization term that is designed to encode the relatedness information between MRTs and $r>0$ is the regularization parameter.

Alternatively, and as mentioned by \citet{kakade2008complexity}, the above regularized optimization can be equivalently rewritten as $$\min\limits_{\mathrm{R}(g^{[1]},\cdots,g^{[M]})\leq c } \sum\limits_{m=1}^M \mathrm{E}^{[m]}_N(\ell^{[m]}\circ g^{[m]}),$$
 where, instead of exploiting the regularization, a hard restriction $\mathrm{R}(g^{[1]},\cdots,g^{[M]})\leq c$ is set to combine the function classes $\mathcal{G}^{[1]},\cdots,\mathcal{G}^{[M]}$, which shrinks the original search space $\bm{\mathcal{G}}$ to $\bm{\mathcal{G}}_c^{\mathrm{R}}$.\footnote{For example, if $g^{[m]}(x^{[m]})=x^{[m]}$ for any $1\leq m\leq M$, the original search space $\bm{\mathcal{G}}$ is the $M$-dimensional real space $\mathbb{R}^M$. Then, by setting the restriction $\sum_{m=1}^M(x^{[m]})^2\leq c^2$, the original space $\bm{\mathcal{G}}$ will become an $M$-dimensional sphere $\bm{\mathcal{G}}_c^{\mathrm{R}}$ with radius $c$.} Therefore, a proper regularization term $\mathrm{R}({\bf g})$ can correctly encode the relatedness between MRTs, reduce the computational cost, and improve the generalization performance. However, this design relies on a prior knowledge of the MRTs.

%\begin{equation}\label{eq:optimization3}
  %\min_{\mathrm{R}(g^{[1]},\cdots,g^{[M]})\leq c } \sum_{m=1}^M \mathrm{E}^{[m]}_N(\ell^{[m]}\circ g^{[m]}),
%\end{equation}

From the vector-valued function learning perspective \citep{micchelli2004kernels,micchelli2005learning}, RMTL aims to find a vector-valued function
${\bf g}_N=(g_N^{[1]},\cdots,g_N^{[M]})^T$ by simultaneously solving the $M$ optimization problems:
\begin{equation}\label{eq:optimization1}
  \min_{{\bf g}\in\bm{\mathcal{G}}_c^{\mathrm{R}}}\left\{ \mathrm{E}^{[m]}_N(\ell^{[m]}\circ g^{[m]}),\;\; 1 \leq m\leq M \right\},
\end{equation}
where $\min\limits_{{\bf g}\in\bm{\mathcal{G}}_c^{\mathrm{R}}}$ stands for a component-wise minimum operator defined in Section \ref{sec:notation}.

%%%%%%%%%%%%%%%%%%%%%%%%%%%%%%%%%%%%
 %
%
%\begin{figure}[htbp]%%%%
%\begin{center}
%\includegraphics[height=3cm]{reshape.eps}
%\caption{{\small The original function class $\bm{\mathcal{G}}$ shrinks to be a new class $\bm{\mathcal{G}}_c^{\mathrm{R}}$ by imposing the hard restriction $\mathrm{R}({\bf g})\leq c$. For example, if $g^{[m]}(x^{[m]})=x^{[m]}$ ($1\leq m\leq M$), the class $\bm{\mathcal{G}}$ is the $M$-dimensional real space $\mathbb{R}^M$ and will become an $M$-dimensional sphere with the radius of $c$ by imposing the restriction $\sum_{m=1}^M(x^{[m]})^2\leq c^2$.}}
%\label{fig:reshape}
%\end{center}
%\end{figure}%%%%

\subsection{Notations of Vector Operations}\label{sec:notation}

For the discussion that follows, it is first necessary to describe some notations of vector operations. Given two vectors, ${\bf x}=(x^{[1]},\cdots,x^{[M]})^T$ and ${\bf y}=(y^{[1]},\cdots,y^{[M]})^T$,
let $|{\bf x}|:=(|x^{[1]}|,\cdots,|x^{[M]}|)^T$ and denote  the expression ${\bf x}>{\bf y}$ (resp. ${\bf x}\geq{\bf y}$) as $x^{[m]}>y^{[m]}\;\;\mbox{(resp. $x^{[m]}\geq y^{[m]}$)}$ for any $1\leq m\leq M$. Similarly, we denote ${\bf x}<{\bf y}$ (resp. ${\bf x}\leq{\bf y}$) as $x^{[m]}< y^{[m]}$ (resp. $x^{[m]}\leq y^{[m]}$) for any $1\leq m\leq M$.

Furthermore, given $(a^{[1]},\cdots,a^{[M]})^T\in\mathbb{R}^M$, we define the component-wise supremum operator
$$\sup\limits_{{\bf g}\in\bm{\mathcal{G}}}\left\{ (g^{[1]}(a^{[1]}),\cdots,g^{[M]}(a^{[M]}))^T\right\}$$
with ${\bf g}=(g^{[1]},\cdots,g^{[M]})^T$ as follows: if the vector-valued function ${\bf g}_\dag=(g_\dag^{[1]},\cdots,g_\dag^{[M]})^T$ achieves the supremum over $\bm{\mathcal{G}}$, each component $g_\dag^{[m]}$ of the vector ${\bf g}_\dag$ achieves the supremum $\sup\limits_{g^{[m]}\in\mathcal{G}^{[m]}}\{g^{[m]}(a^{[m]})\}$ over $\mathcal{G}^{[m]}$.
Similarly, we define the component-wise minimum operator as $$\min\limits_{{\bf g}\in\bm{\mathcal{G}}}\left\{ (g^{[1]}(a^{[1]}),\cdots,g^{[M]}(a^{[M]}))^T\right\}.$$

\subsection{Task-joint Probability and Generalization Bounds}

%\begin{equation}\label{eq:supremum}
 %\sup_{{\bf g}\in\bm{\mathcal{G}}}\left\{ (g^{[1]}(a^{[1]}),\cdots,g^{[M]}(a^{[M]}))^T\right\}\quad\mbox{with\;\; ${\bf g}=(g^{[1]},\cdots,g^{[M]})^T$}
%\end{equation}
 %is defined as follows: if the vector-valued function ${\bf g}_\dag=(g_\dag^{[1]},\cdots,g_\dag^{[M]})^T$ achieves the supremum \eqref{eq:supremum} w.r.t. $(a^{[1]},\cdots,a^{[M]})^T$ over $\bm{\mathcal{G}}$, each component $g_\dag^{[m]}$  ($1\leq m\leq M$) of the vector ${\bf g}_\dag$ achieves the supremum $\sup_{g^{[m]}\in\mathcal{G}^{[m]}}\{g^{[m]}(a^{[m]})\}$ w.r.t. $a^{[m]}$ over $\mathcal{G}^{[m]}$.
%Similarly, we define the component-wise minimum operator as
%\begin{equation}\label{eq:minimum}
 %\min_{{\bf g}\in\bm{\mathcal{G}}}\left\{ (g^{[1]}(a^{[1]}),\cdots,g^{[M]}(a^{[M]}))^T\right\}.
 %\end{equation}

In general, the generalization bounds for STL refer to the upper bounds of the supremum  $$\sup\limits_{g\in\mathcal{G}}|\mathrm{E}(\ell\circ g)-\mathrm{E}_N(\ell\circ g)|$$
with an alternative probability expression $$\mathrm{Pr}\big\{\sup\limits_{g\in\mathcal{G}}|\mathrm{E}(\ell\circ g)-\mathrm{E}_N(\ell\circ g)|>\xi\big\},$$ whose upper bound describes the rarity of the event that the {\it empirical discrepancy} between the expected risk $\mathrm{E}(\ell\circ g)$ and the empirical risk $\mathrm{E}_N(\ell\circ g)$ is larger than a given positive constant $\xi$.

%\begin{equation}\label{eq:sbound1}
  %\sup_{g\in\mathcal{G}}|\mathrm{E}(\ell\circ g)-\mathrm{E}_N(\ell\circ g)|.
%\end{equation}
%Alternatively, the generalization bounds \eqref{eq:sbound1} can also be represented in the probability form \citep[see][]{Bousquet04}:
%\begin{equation}\label{eq:sbound2}
 % \mathrm{Pr}\left\{\sup_{g\in\mathcal{G}}|\mathrm{E}(\ell\circ g)-\mathrm{E}_N(\ell\circ g)|>\xi\right\},\end{equation}
%whose upper bound describes the probability of the event that the discrepancy between the expected risk $\mathrm{E}(\ell\circ g)$ and the empirical risk $\mathrm{E}_N(\ell\circ g)$ is larger than a given constant $\xi>0$.

%As addressed by \citet{Bousquet04,zhang2013bennett}

Since MRTs are processed simultaneously in MTL, the following task-joint probability is straightforward: for any $\bm{\xi}=(\xi^{[1]},\cdots,\xi^{[M]})^T>0$,
\begin{equation}\label{eq:joint.prob}
 \mathrm{Pr}\left\{
                      \sup_{{\bf g}\in\bm{\mathcal{G}}_c^{\mathrm{R}}}\left\{\begin{pmatrix}
                                  |\mathrm{E}^{[1]}(\ell^{[1]}\circ g^{[1]})-\mathrm{E}^{[1]}_N(\ell^{[1]}\circ g^{[1]}) |\\
                                  \vdots \\
                                   |\mathrm{E}^{[M]}(\ell^{[M]}\circ g^{[M]})-\mathrm{E}^{[M]}_N(\ell^{[M]}\circ g^{[M]})|\\
                                \end{pmatrix}\right\}\not\leq\begin{pmatrix}
                                  \xi^{[1]}\\
                                  \vdots \\
                                   \xi^{[M]}\\
                                \end{pmatrix}\right\},
\end{equation}
which describes the rarity of the event in RMTL that there is at least one task $\mathcal{Z}^{[m]}$ with empirical discrepancy larger than the constant $\xi^{[m]}$. The upper bound of \eqref{eq:joint.prob} is the so-called ``generalization bound" for RMTL.
Compared to the STL bound, the RMTL bound \eqref{eq:joint.prob} not only reflects the generalization performance of each task, but also the dependence between simultaneously learned tasks, {\it i.e.,} how the success (or failure) of some tasks affects the performance of the others.

For convenience, we further define the loss function
class:
\begin{equation}\label{eq:fclass.1}
    \mathcal{F}^{[m]}:=\{{\bf z}^{[m]} \mapsto \ell^{[m]}(g^{[m]}({\bf x}^{[m]}),{\bf y}^{[m]}):g^{[m]}\in \mathcal{G}^{[m]}\},\;\;1\leq m\leq M;
\end{equation}
the Cartesian product $\bm{\mathcal{F}}:=\mathcal{F}^{[1]}\times\cdots\times\mathcal{F}^{[M]}$ is called the ``vector-valued function class" in the rest of this paper. Similarly, based on the regularized vector-valued function class $\bm{\mathcal{G}}_c^\mathrm{R}$, we define the regularized loss vector-valued function class by
\begin{equation}\label{eq:fclass.2}
    \bm{\mathcal{F}}_c^\mathrm{R}:=\left\{(\ell^{[1]}\circ g^{[1]},\cdots,\ell^{[M]}\circ g^{[M]})^T
     :(g^{[1]},\cdots,g^{[M]})^T\in \bm{\mathcal{G}}_c^\mathrm{R}\right\},
\end{equation}
which is also termed the regularized vector-valued function class in the remainder of this paper.
Briefly, we denote for any ${\bf f}:=(f^{[1]},\cdots,f^{[M]})^T\in\bm{\mathcal{F}}$,
\begin{equation}\label{eq:short1}
    \mathrm{E}^{[m]}f^{[m]}:=\int f^{[m]}({\bf z}^{[m]})d\mathrm{P}^{[m]}({\bf z}^{[m]})\;\;;\;\;\mathrm{E}^{[m]}_Nf^{[m]}:= \frac{1}{N}\sum_{n=1}^{N}f^{[m]}({\bf z}_n^{[m]}),
\end{equation}
and the generalization bound \eqref{eq:joint.prob} is equivalently rewritten as $\mathrm{Pr}\big\{\sup\limits_{{\bf f} \in\bm{\mathcal{F}}_c^{\mathrm{R}}}
\big\{|{\bf E}{\bf f}-{\bf E}_N{\bf f}|\}\not\leq\bm{\xi}\big\}
$
with $${\bf E}{\bf f}:=(\mathrm{E}^{[1]}f^{[1]},\cdots,\mathrm{E}^{[M]}f^{[M]})^T$$ and $${\bf E}_N{\bf f}:=(\mathrm{E}_N^{[1]}f^{[1]},\cdots,\mathrm{E}_N^{[M]}f^{[M]})^T.$$

%%
%\begin{equation}\label{eq:mbound2}
%\mathrm{Pr}\left\{\sup_{{\bf f} \in\bm{\mathcal{F}}_c^{\mathrm{R}}}
%\big\{|{\bf E}{\bf f}-{\bf E}_N{\bf f}|\big\}>\bm{\xi}\right\}
%\end{equation}

\section{Measures of Task-group Relatedness}\label{sec:measure}

Some existing works on task relatedness already describe the relationship between two individual tasks, for instance the $\mathcal{F}$-related \citep{ben2003exploiting,ben2008notion} notion and covariances \citep{Zhang10convex}. In MTL, it is also necessary to consider the relationship between two task groups. Here, we present two measures of task-group relatedness: the observed discrepancy-dependence measure (ODDM) and the empirical discrepancy-dependence measure (EDDM).

%Moreover, we also present an extension of covariances (ExCov), which is of individual type, to highlight the necessity of the negative correlation in the multi-task learning, which is in accordance with the argument in \citep{Zhang10convex}.

%To our best knowledge, there is no theoretical result on the analytical relationship between the covariance matrix and the generalization performance of the multi-task learning.

\subsection{ODDM }

%Here, we will introduce a novel quantity to measure the relatedness among multiple tasks.
 %\begin{equation}\label{eq:dependence}
%\mathrm{Pr}\{A|B\}-\mathrm{Pr}\{A\},
%\end{equation}

In probability theory, the dependence between two events $\mathcal{A}$ and $\mathcal{B}$ can be detected using the quantity $\mathrm{Pr}\{\mathcal{A}|\mathcal{B}\}-\mathrm{Pr}\{\mathcal{A}\}$,
where $\mathcal{A}$ and $\mathcal{B}$ are positively dependent if the conditional probability $\mathrm{Pr}\{\mathcal{A}|\mathcal{B}\}$ of $\mathcal{A}$ given $\mathcal{B}$ is greater than the probability $\mathrm{Pr}\{\mathcal{A}\}$ ({\it i.e.}, $\mathrm{Pr}\{\mathcal{A}|\mathcal{B}\}-\mathrm{Pr}\{\mathcal{A}\}>0$), and they are negatively dependent if the inequality is reversed \citep{bradley2005basic,mohri2010stability}. We introduce ODDM and EDDM to measure the relatedness between two task groups in MTL, based on the quantity.

\begin{definition}\label{def:ODDM}
Given $M$ tasks $\mathcal{Z}^{[1]},\cdots,\mathcal{Z}^{[M]}$ and a regularized vector-valued function class $\bm{\mathcal{F}}_c^{\mathrm{R}}$,
let $\bm{\Lambda}:=\{1,\cdots,M\}$ be an index set and $\Lambda^{[m]}$ be a subset of $\bm{\Lambda}$ with the cardinality of $m$. For any $\Lambda^{[m]}\subset\bm{\Lambda}$ and any $\bm{\xi}=(\xi^{[1]},\cdots,\xi^{[M]})^T>{\bf 0}$, ODDM is defined as
\begin{align*}
\phi_{\bm{\mathcal{F}}}(\Lambda^{[m]},\bm{\xi}):=\sup_{{\bf f}\in\bm{\mathcal{F}}_c^{\mathrm{R}}}
\Big\{ \mathrm{Pr}\big\{  \mathcal{A}_{\Lambda^{[m]}} \big|
\mathcal{B}_{\Lambda^{[m]}}  \big\}-\mathrm{Pr}\big\{  \mathcal{A}_{\Lambda^{[m]}}   \big\}
\Big\},
\end{align*}
where ${\bf f}=(f^{[1]},\cdots,f^{[M]})^T$, $\overline{\Lambda^{[m]}}$ stands for the complementary set of $\Lambda^{[m]}$ with $\Lambda^{[m]}\cup\overline{\Lambda^{[m]}}=\bm{\Lambda}$, and the events $\mathcal{A}_{\Lambda^{[m]}} := \{s^{[i]}>\xi^{[i]}\}_{i\in\Lambda^{[m]}}$ and $\mathcal{B}_{\Lambda^{[m]}} :=\{s^{[i]}\leq\xi^{[i]}\}_{i\in\overline{\Lambda^{[m]}}}$ w.r.t. the {\it observed discrepancy} $$s^{[i]}:=|\mathrm{E}^{[i]}f^{[i]}-f^{[i]}({\bf z}^{[i]})|$$ of the task $\mathcal{Z}^{[m]}$.
\end{definition}

As defined above, ODDM measures the dependence between the events that the tasks in group $\Lambda^{[m]}$ have large observed discrepancies and the tasks in $\overline{\Lambda^{[m]}}$ have small observed discrepancies.
In fact, ODDM is determined by the inherent characteristics of MRTs, the selection of function classes and the regularization term.
It can exist in one of three states:
\begin{itemize}
  \item a positive ODDM implies that some functions in the search space $\bm{\mathcal{F}}_c^{\mathrm{R}}$ will result in a negative synergy effect between the tasks $\{\mathcal{Z}^{[i]}\}_{i\in\overline{\Lambda^{[m]}}}$ and the others $\{\mathcal{Z}^{[i]}\}_{i\in\Lambda^{[m]}}$, {\it i.e.,} the success of tasks $\{\mathcal{Z}^{[i]}\}_{i\in\overline{\Lambda^{[m]}}}$ will benefit from a performance loss in the others $\{\mathcal{Z}^{[i]}\}_{i\in\Lambda^{[m]}}$;\vspace{-2mm}
  \item a negative ODDM means that all functions in $\bm{\mathcal{F}}_c^{\mathrm{R}}$ will effect the synergy effect on the simultaneous learning process for MRTs, {\it i.e.,} the success of the tasks $\{\mathcal{Z}^{[i]}\}_{i\in\overline{\Lambda^{[m]}}}$ contributes to improved performance of the others $\{\mathcal{Z}^{[i]}\}_{i\in\Lambda^{[m]}}$;\vspace{-2mm}
  \item a zero ODDM reflects that some functions in $\bm{\mathcal{F}}_c^{\mathrm{R}}$ eliminate the relatedness between  $\{\mathcal{Z}^{[i]}\}_{i\in\Lambda^{[m]}}$ and $\{\mathcal{Z}^{[i]}\}_{i\in\overline{\Lambda^{[m]}}}$, and the others will effect synergy effect between the two groups.
\end{itemize}

%

%
%Actually, ODDM is determined by the inherent characteristics of MRTs, the selection of function classes and the regularization term, and thus MTL will benefit a lot from a good regularization term $\mathrm{R}(\bm{\mathcal{F}})$, which is in accordance with many existing empirical evidences.

\subsection{EDDM}

Since this paper focuses on ERM-based RMTL, we also need to consider the asymptotic behavior of the dependence between two task groups when the sample size $N$ goes to {\it infinity}.
%As shown in \eqref{eq:mbound}, the vector-valued generalization bound aims to describe the synergism among the multiple tasks in the multi-task learning, {\it i.e.,} how the performance of each task is affected by the other tasks in the simultaneous learning process.
\begin{definition}\label{def:EDDM}
Following the notations in Definition \ref{def:ODDM} and letting ${\bf Z}_{N}^{[m]}:=\{{\bf z}^{[m]}_n\}_{n=1}^{N}$ be $N$ i.i.d. samples drawn from each task $\mathcal{Z}^{[m]}$ ($1\leq m\leq M$), EDDM is defined as
\begin{align*}
\varphi^N_{\bm{\mathcal{F}}_c^{\mathrm{R}}}(\Lambda^{[m]},\bm{\xi})
:=
\mathrm{Pr}\big\{  \mathcal{A}^N_{\Lambda^{[m]}} \big|
\mathcal{B}^N_{\Lambda^{[m]}}  \big\}-\mathrm{Pr}\big\{ \mathcal{A}^N_{\Lambda^{[m]}} \big\},
\end{align*}
where the events $\mathcal{A}^N_{\Lambda^{[m]}}:= \{t_N^{[i]}>\xi^{[i]}\}_{i\in\Lambda^{[m]}}$ and $\mathcal{B}^N_{\Lambda^{[m]}}:=\{t_N^{[i]}\leq\xi^{[i]}\}_{i\in\overline{\Lambda^{[m]}}} $ with the {\it empirical discrepancy}
\begin{equation}\label{eq:t}
  t_N^{[i]}:=\sup\limits_{f\in\mathrm{Prj^{[i]}}
(\bm{\mathcal{F}}_c^{\mathrm{R}})}|\mathrm{E}^{[i]}f-\mathrm{E}_N^{[i]}f|,
\end{equation}
w.r.t. the sample set ${\bf Z}_{N}^{[m]}$ drawn from $\mathcal{Z}^{[m]}$, and $\mathrm{Prj^{[i]}}(\bm{\mathcal{F}}_c^{\mathrm{R}})$ stands for the projection of the regularized vector-valued function class $\bm{\mathcal{F}}_c^{\mathrm{R}}$ onto the function class $\mathcal{F}^{[i]}$.
\end{definition}
Note that EDDM measures the dependence between the generalization performances of the two task groups and also has three states:
\begin{itemize}
  \item a positive EDDM implies that the successfully learned tasks $\{\mathcal{Z}^{[i]}\}_{i\in\overline{\Lambda^{[m]}}}$ benefit from a loss in generalization performance of the others $\{\mathcal{Z}^{[i]}\}_{i\in\Lambda^{[m]}}$;
  \item a negative EDDM means that the task groups $\{\mathcal{Z}^{[i]}\}_{i\in\Lambda^{[m]}}$ and $\{\mathcal{Z}^{[i]}\}_{i\in\overline{\Lambda^{[m]}}}$ are mutually beneficial;
  \item a zero EDDM with $N<\infty$ signifies that the two groups are unrelated.
\end{itemize}

\subsection{Empirically Computing ODDM and EDDM}\label{sup:compute}

By the facts that $\mathrm{Pr}\{\mathcal{A}|\mathcal{B}\}=\mathrm{Pr}\{\mathcal{A},\mathcal{B}\}/\mathrm{Pr}\{\mathcal{B}\}$ and $\mathrm{Pr}\{\mathcal{A}\}=\mathrm{E}\mathbf{1}_{\{\mathcal{A}\}}$, ODDM $\phi_{\bm{\mathcal{F}}}(\Lambda^{[m]},\bm{\xi})$ can be empirically computed in the following way. Letting $\{{\bf z}^{[m]}_n\}_{n=1}^{N}$ be i.i.d. samples drawn from the task $\mathcal{Z}^{[m]}$ ($1\leq m\leq M$), we denote $\zeta_j$ ($1\leq j\leq J$), $\eta_k$ ($1\leq k\leq K$) and $\theta_p$ ($1\leq p\leq P$) as the observations of the events $\mathcal{A}_{\Lambda^{[m]}} \wedge\mathcal{B}_{\Lambda^{[m]}} $, $\mathcal{A}_{\Lambda^{[m]}} $ and $\mathcal{B}_{\Lambda^{[m]}}$, respectively. Then, an empirical version of ODDM $\phi_{\bm{\mathcal{F}}}(\Lambda^{[m]},\bm{\xi})$ is given by:
\begin{align}\label{eq:ODDM.1}
\widehat{\phi}_{\bm{\mathcal{F}}}(\Lambda^{[m]},\bm{\xi}):=\sup_{{\bf f}\in\bm{\mathcal{F}}_c^{\mathrm{R}}}
\bigg\{ \frac{J^{-1}\sum_{j=1}^{J}\mathbf{1}_{\{  \zeta_j \}}}{P^{-1}\sum_{p=1}^{P}\mathbf{1}_{\{\theta_p  \}}}-K^{-1}\sum_{k=1}^{K}\mathbf{1}_{
  \{\eta_k\}}
\bigg\},
\end{align}
where the expected risk $\mathrm{E}^{[i]}f^{[i]}$ in $s^{[i]}$ is approximated by its empirical version $\mathrm{E}_N^{[i]}f^{[i]}$.

Recalling the term $t_N^{[i]}$ defined in \eqref{eq:t}, EDDM $\varphi^N_{\bm{\mathcal{F}}_c^{\mathrm{R}}}(\Lambda^{[m]},\bm{\xi})$ can be approximately computed in the following way. First, fix the sample set $\{{\bf z}^{[i]}_n\}_{n=1}^{N}$ of each task $\mathcal{Z}^{[i]}$ ($1\leq i\leq M$) and replace the expected risk $\mathrm{E}^{[i]}f$ with the fixed empirical quantity $\mathrm{E}_{N}^{[i]}f$ w.r.t. $\{{\bf z}^{[i]}_n\}_{n=1}^{N}$.
Next, we randomly select $L$ samples from of each task $\mathcal{Z}^{[i]}$ to form another empirical risk $\mathrm{E}_L^{[i]}f$ and denote $\hat{t}_{L}^{[i]}:=\sup\limits_{f\in\mathrm{Prj^{[i]}}
(\bm{\mathcal{F}}_c^{\mathrm{R}})}|\mathrm{E}_{L}^{[i]}f-\mathrm{E}_N^{[i]}f|$ as an estimate of $t_N^{[i]}$. Denote the events $\mathcal{A}_L:= \{\hat{t}_{L}^{[i]}>\xi^{[i]}\}_{i\in\Lambda^{[m]}}$ and $\mathcal{B}_L:=\{\hat{t}_{L}^{[i]}\leq\xi^{[i]}\}_{i\in\overline{\Lambda^{[m]}}}$.
Let $\zeta_j$ ($1\leq j\leq J$), $\eta_k$ ($1\leq k\leq K$) and $\theta_p$ ($1\leq p\leq P$) be the observations of the events $\mathcal{A}_{L} \wedge\mathcal{B}_{L} $, $\mathcal{A}_{L} $ and $\mathcal{B}_{L}$ respectively. We then can empirically compute EDDM $\varphi^N_{\bm{\mathcal{F}}_c^{\mathrm{R}}}(\Lambda^{[m]},\bm{\xi})$ as \vspace{-2mm}
\begin{align}\label{eq:EDDM}\vspace{-2mm}
\widehat{\varphi}^N_{\bm{\mathcal{F}}_c^{\mathrm{R}}}(\Lambda^{[m]},\bm{\xi}):=  \frac{J^{-1}\sum_{j=1}^{J}\mathbf{1}_{\{  \zeta_j \}}}{P^{-1}\sum_{p=1}^{P}\mathbf{1}_{\{\theta_p  \}}}-K^{-1}\sum_{k=1}^{K}\mathbf{1}_{
  \{\eta_k\}}.\vspace{-2mm}
\end{align}
\begin{remark}
There are two difficulties to implement this method to empirically compute ODDM and EDDM:
\begin{itemize}
  \item In general, it is hard to capture the observations of the task-joint events.
  \item If the task number $M$ is large, it is highly time-consuming to compute the empirical estimates of ODDM and EDDM for any $\Lambda^{[m]}$. To reduce the complexity, one feasible way is to cluster the tasks according to the similarity and select a representative task from each cluster to compute ODDM and EDDM.
\end{itemize}
\end{remark}

%-----------------------------------------------------------------

\section{Cartesian Product-based Uniform Entropy Numbers}\label{sec:UEN}

Complexity measures of function classes play an important role in learning theory. Since this paper studies MTL in the vector-valued framework, the classical measures such as the Vapnik-Chervonenkis (VC) dimension and the covering number, are not applicable (or at least cannot be directly applied) to the vector-valued scenario. For example, \citet{ben2008notion} applied an extended version of the VC dimension to study the generalization properties of multi-task classification.

Here, we introduce the Cartesian product-based uniform entropy number (CPUEN) to measure the complexity of the vector-valued function classes. First, we briefly outline the definitions of the covering number and uniform entropy number (UEN) of the scalar-valued function classes. Regarding further details, please refer to \citet{Mendelson03}.

\begin{definition}\label{def:CovNum}
Let $\mathcal{F}$ be a function class and $d$ be a metric on $\mathcal{F}$. For any $\xi>0$, the covering number of $\mathcal{F}$ at radius $\xi$
w.r.t. the metric $d$, denoted by $\mathcal{N}(\mathcal{F},\xi,d)$ is the minimum size of a cover of radius $\xi$. Furthermore, given a sample set ${\bf Z}_{N}:=\{{\bf z}_{n}\}_{n=1}^{N}$ drawn from $\mathcal{Z}$, we denote ${\bf Z}'_{N}:=\{{\bf z}'_{n}\}_{n=1}^{N}$ as the ghost sample set drawn from $\mathcal{Z}$, such that the ghost sample ${\bf z}'_{n}$ has the same distribution as ${\bf z}_{n}$ for any $1\leq n\leq N$.
Denote ${\bf Z}_{2N}:=\{{\bf Z}_{N},{\bf Z}'_{N}\}$. Setting the metric $d$ as the $\ell_p({\bf Z}_{2N})$ ($p>0$) norm, UEN is defined by
\begin{equation}\label{eq:UEN0}
 \ln\mathcal{N}_p(\mathcal{F},\xi,N):=\sup\limits_{{\bf Z}_{N}}\ln\mathcal{N}\left(\mathcal{F},\xi,\ell_p({\bf Z}_{N})\right).
\end{equation}
\end{definition}

Recall that the vector-valued function class $\bm{\mathcal{F}}$ is a Cartesian product of the function classes $\mathcal{F}^{[1]},\cdots,\mathcal{F}^{[M]}$, {\it i.e.,}
$\bm{\mathcal{F}}:=\mathcal{F}^{[1]}\times\cdots\times\mathcal{F}^{[M]}$. For each $\mathcal{F}^{[m]}$ ($1\leq m\leq M$), let $\widetilde{{\bf Z}}^{[m]}_{N}$ be the sample set achieving the supremum \begin{equation}\label{eq:UEN2}
   \sup_{{\bf Z}^{[m]}_{N}\in(\mathcal{Z}^{[m]})^{N}}\ln\mathcal{N}\left(\mathcal{F}^{[m]},\xi^{[m]},\ell_p({\bf Z}^{[m]}_{N})\right)
 \end{equation}
 and $\Omega_{p,N}^{[m]}$ be one of the covers of $\mathcal{F}^{[m]}$ related to the supremum w.r.t. the norm $\ell_p(\widetilde{{\bf Z}}^{[m]}_{N})$. Therefore, the Cartesian product $\Omega_{p,N}^{[1]}\times\cdots
\times\Omega_{p,N}^{[M]}$ is also a cover of $\bm{\mathcal{F}}$ with the radius vector $\bm{\xi}:=(\xi^{[1]},\cdots,\xi^{[M]})^T$. Following the above notations,
we define the CPUEN of the vector-valued function class $\bm{\mathcal{F}}$ as follows:

\begin{definition}\label{def:UEN}
Given a vector-valued function class $\bm{\mathcal{F}}$, consider a Cartesian product-based cover of the vector-valued function $\bm{\mathcal{F}}$: $$\bm{\Omega}_{p,N}(\bm{\mathcal{F}},\bm{\xi}):=\Big\{\bm{\mathcal{A}_p^M}
  \in\Omega_{p,N}^{[1]}\times
  \cdots\times\Omega_{p,N}^{[M]}:\bm{\mathcal{A}_p^M}\cap
  \bm{\mathcal{F}}\not=\varnothing\Big\}.$$ Then, CPUEN of $\bm{\mathcal{F}}$ is defined as
$\ln\bm{\mathcal{N}}_p(\bm{\mathcal{F}},\bm{\xi},N):=\ln |\bm{\Omega}_{p,N}(\bm{\mathcal{F},\bm{\xi}})|$.
\end{definition}

%\begin{equation}\label{eq:cover}
  %\bm{\Omega}_{p,N}(\bm{\mathcal{F}},\bm{\xi}):=\Big\{\bm{\mathcal{A}_2^M}
  %\in\Omega_{p,N}^{[1]}\times
  %\cdots\times\Omega_{p,N}^{[M]}:\bm{\mathcal{A}_2^M}\cap
  %\bm{\mathcal{F}}\not=\varnothing\Big\}.
%\end{equation}
%\begin{equation}\label{eq:UEN}
  %\ln\bm{\mathcal{N}}_p(\bm{\mathcal{F}},\bm{\xi},N):=\ln |\bm{\Omega}_{p,N}(\bm{\mathcal{F},\bm{\xi}})|.
%\end{equation}
%
In contrast to the classical UEN [see \eqref{eq:UEN0}], CPUEN is induced from the cover of the function class $\mathcal{F}^{[m]}$ of each task $\mathcal{Z}^{[m]}$ ($1\leq m\leq M$) with different norms and radiuses instead of introducing a uniform norm in the vector-valued function space $\bm{\mathcal{F}}$. Although CPUEN is usually larger than the uniform-norm UEN of the vector-valued function class $\bm{\mathcal{F}}$, the induction setting of CPUEN has a stronger relationship with the prior information-based design of the regularization term and offers convenience to the theoretical analysis of RMTL.

%--------------------------------------------------------

\section{Generalization Bounds of Regularized Multi-task Learning}\label{sec:main}

In this section, we present the generalization bounds of RMTL and discuss how the task-group relatedness affects the generalization properties of RMTL. Moreover, we give a sufficient condition for the consistency of each task in MRTs.

\subsection{Two Special Cases}

Before the formal discussion, we first bound the probabilities of two special events: first, that all tasks have large empirical discrepancies and second, that all tasks have small empirical discrepancies.

%\begin{equation}\label{eq:sym.condition.1.A}
%   N\geq \frac{8\Gamma(\bm{\Lambda})}{1-2\Upsilon(\bm{\Lambda})},
%\end{equation}

\begin{theorem}\label{thm:bound.A}
Assume that $\bm{\mathcal{F}}_c^{\mathrm{R}}$ is a regularized vector-valued function class w.r.t. the constant $c$, and ${\bf Z}^{[m]}_N=\{{\bf z}^{[m]}_{n}\}_{n=1}^{N}$ is the set of $N$ i.i.d. samples drawn from the task $\mathcal{Z}^{[m]}$ ($1\leq m\leq M$). Let $\bm{\Lambda}:=\{1,\cdots,M\}$ be an index set and denote $\Lambda^{[m]}$ as a subset of $\bm{\Lambda}$ with the cardinality of $m$. Denote ${\bf Z}^{[m]}_{2N}:=\{{\bf Z}^{[m]}_N,{\bf Z'}^{[m]}_N \}$. Given $\bm{\xi}=(\xi^{[1]},\cdots,\xi^{[M]})^T>{\bf 0}$ and for any $N\in\mathbb{N}$ such that $N\geq \frac{8\Gamma(\bm{\Lambda})}{1-2\Upsilon(\bm{\Lambda})}$,
it then holds that
\begin{align}\label{eq:bound.A}
  \mathrm{Pr}\left\{\sup_{{\bf f} \in\bm{\mathcal{F}}_c^\mathrm{R}}
\big|{\bf E}{\bf f}-{\bf E}_N{\bf f}\big|> {\bm \xi} \right\}\leq 2^{M+2} \bm{\mathcal{N}}_1\big(\bm{\mathcal{F}}_c^{\mathrm{R}},\bm{\xi}/8,2N\big) \exp\left\{\frac{-N\sum_{m=1}^M(\xi^{[m]})^2}{32M^2(b-a)^2}\right\},
\end{align}
where
\begin{equation}\label{eq:Gamma0}
  \Gamma(\bm{\Lambda}):=\sum_{m=1}^M\sum_{\Lambda^{[m]}\subset\bm{\Lambda}}\frac{m(b-a)^2}{\sum\limits_{i\in \Lambda^{[m]}}(\xi^{[i]})^2},
\end{equation}
and
\begin{equation}\label{eq:eq:Upsilon0}
  \Upsilon(\bm{\Lambda}):=\sum_{m=1}^M\sum_{\Lambda^{[m]}\subset\bm{\Lambda}}
  \phi_{\bm{\mathcal{F}}}(\Lambda^{[m]},\bm{\xi}).
\end{equation}
\end{theorem}

This theorem shows that if it holds that $N\geq \frac{8\Gamma(\bm{\Lambda})}{1-2\Upsilon(\bm{\Lambda})}$, the probability of $\sup\limits_{{\bf f} \in\bm{\mathcal{F}}_c^\mathrm{R}}
\big|{\bf E}{\bf f}-{\bf E}_N{\bf f}\big|> {\bm \xi}$ can be bounded by the RHS of \eqref{eq:bound.A}. Note that if $M=1$, since $\phi_{\bm{\mathcal{F}}}(\Lambda^{[m]},\bm{\xi})$ equals {\it zero}, the quantity $\Upsilon(\bm{\Lambda})$ is {\it zero} and the bound \eqref{eq:bound.A} coincides with the classical result of STL (see Theorem 2.3 of \citep{Mendelson03}).

\begin{remark}\label{rem:ODDM1}
In the case of $M>1$, the condition $N\geq \frac{8\Gamma(\bm{\Lambda})}{1-2\Upsilon(\bm{\Lambda})}$ should be satisfied when the quantity $\Upsilon(\bm{\Lambda})<1/2$: namely, it is necessary for RMTL to satisfy the condition that the task-group relatedness between MRTs should mostly be synergistic. Furthermore, RMTL will perform well with less samples than STL size $N\geq\frac{8(b-a)^2}{\xi^2}$ if the condition $\Upsilon(\bm{\Lambda})\leq (1-2^{M-1})$ holds, which implies that almost any pair of task groups $\Lambda^{[m]}$ and $\overline{\Lambda^{[m]}}$ predominantly promote mutually.\footnote{Actually, letting $\xi_0:=\min\{\xi^{[1]},\cdots,\xi^{[M]}\}$ and $N_0:=\frac{8(b-a)^2}{\xi_0^2}$, we have $8\Gamma(\bm{\Lambda})<(2^M-1)N_0$. Thus, the condition $N\geq \frac{8\Gamma(\bm{\Lambda})}{1-2\Upsilon(\bm{\Lambda})}$ holds if $N$ is larger than $\frac{(2^M-1)N_0}{1-2\Upsilon(\bm{\Lambda})}$. We can then infer that each task in RMTL will need less samples than the task in STL if the condition $\frac{2^M-1}{1-2\Upsilon(\bm{\Lambda})}<1$ holds. }

\end{remark}

\begin{remark}\label{rem:ODDM2}
If $\xi=\xi^{[1]}=\cdots=\xi^{[M]}$ and each ODDM $\phi_{\bm{\mathcal{F}}}(\Lambda^{[m]},\bm{\xi})$ reaches the minimum value $-1$, the sample size $N$ of each task should be larger than the value $\frac{8(b-a)^2}{((2^M-1)^{-1}+2)\xi^2}$ ($M>1$) to support the inequality \eqref{eq:bound.A}. This implies that the required sample size of each task in RMTL will approach half of the STL value $8(b-a)^2/\xi^2$ at the rate of $2^{-M}$ as $M\rightarrow\infty$. This finding shows that if the relationship between any pair of task groups $\Lambda^{[m]}$ and $\overline{\Lambda^{[m]}}$ is predominantly synergistic, each task in RMTL needs less samples than STL and the required sample size $N$ in RMTL will not increase dramatically, regardless of a large number of MRTs.
\end{remark}

We next consider the second special case and
present an upper bound of the probability that all tasks have small empirical discrepancies in the simultaneous learning process for MRTs. The following theorem is proved by using the small-deviation techniques \citep{Li2012small}.

\begin{theorem}\label{thm:small.prob}
Following the notations in Theorem \ref{thm:bound.A}, it then holds that for any $\bm{\xi}=(\xi^{[1]},\cdots,\xi^{[M]})^T>{\bf 0}$,\vspace{-2mm}
\begin{align}\label{eq:small.prob}
  \mathrm{Pr}\left\{\sup_{{\bf f} \in\bm{\mathcal{F}}_c^\mathrm{R}}
\big|{\bf E}{\bf f}-{\bf E}_N{\bf f}\big|\leq {\bm \xi} \right\}\leq
2^M\sup_{{\bf f} \in\bm{\mathcal{F}}_c^\mathrm{R}}\mathrm{Pr}\big\{
{\bf s}\leq 2{\bm \xi} \big\},\vspace{-2mm}
\end{align}
where ${\bf s}=\big(s^{[1]},\cdots,s^{[M]}\big)^T$ with $s^{[m]}:=\big|\mathrm{E}^{[m]}f^{[m]}
-f^{[m]}({\bf z}^{[m]})\big|$ for any $1\leq m\leq M$.
\end{theorem}

This theorem converts the case of small empirical discrepancies into a simple case, where the LHS of \eqref{eq:small.prob} can be bounded by using the probability that
the observed discrepancy of each task $\mathcal{Z}^{[m]}$ is smaller than $2\xi^{[m]}$ ($1\leq m\leq M$).
Compared to the case of empirical discrepancies, the
RHS of \eqref{eq:small.prob} is only determined by the inherent characteristics of MRTs, {\it e.g.,} the distributions of tasks, the selection of function classes, and the regularization term.

\subsection{Main Results}

Based on these two special cases, we obtain the generalization bounds of RMTL and a sufficient condition for the consistency of each task in the simultaneous learning process for MRTs.

\begin{theorem}\label{thm:main}
Following the notations of Theorem \ref{thm:bound.A}, given $\bm{\xi}=(\xi^{[1]},\cdots,\xi^{[M]})^T>{\bf 0}$ and for any $N\in\mathbb{N}$ such that $N\geq \max\limits_{1\leq m\leq M}\max\limits_{\Lambda^{[m]}\subset\bm{\Lambda}}
\frac{8\Gamma(\Lambda^{[m]})}{1-2\Upsilon(\Lambda^{[m]})}$, it then holds that
\begin{align}\label{eq:main}
 & \mathrm{Pr}\left\{\sup_{{\bf f} \in\bm{\mathcal{F}}_c^\mathrm{R}}
\big|{\bf E}{\bf f}-{\bf E}_N{\bf f}\big|\not\leq {\bm \xi} \right\}
\leq \sum_{m=1}^M\sum_{\Lambda^{[m]}\subset\bm{\Lambda}}
2^m\mathrm{Pr}\left\{ \big\{ s^{[\lambda]}\leq2\xi^{[\lambda]} \big\}_{\lambda\in\overline{\Lambda^{[m]}}}  \right\}\nonumber\\
&\times
\left(\varphi^N_{\bm{\mathcal{F}}_c^{\mathrm{R}}}(\Lambda^{[m]},\bm{\xi})+
2^{m+2} \bm{\mathcal{N}}_1\big(\mathrm{Prj}_{\Lambda^{[m]}}^{\bm{\mathcal{F}}_c^{\mathrm{R}}},\frac{\bm{\xi}_{\Lambda^{[m]}}}{8},2N\big)
\exp\left\{\frac{-N\sum\limits_{\lambda\in\Lambda^{[m]}}(\xi^{[\lambda]})^2}{32M^2(b-a)^2}\right\}   \right),
\end{align}
where $\mathrm{Prj}_{\Lambda^{[m]}}^{\bm{\mathcal{F}}_c^{\mathrm{R}}}$ stands for the projection of $\bm{\mathcal{F}}_c^{\mathrm{R}}$ on the subspace $\prod\limits_{\lambda\in\Lambda^{[m]}}\mathcal{F}^{[\lambda]}$, $\bm{\xi}_{\Lambda^{[m]}}:=\big( \xi^{[\lambda]}\big)_{\lambda\in\Lambda^{[m]}}$, $\Gamma(\Lambda^{[m]})$ and $\Upsilon(\Lambda^{[m]})$ are defined in \eqref{eq:Gamma0}. Furthermore, if it is satisfied that for any $1\leq m\leq M$ and $\lambda^{[m]}\subset\bm{\Lambda}$,
\begin{equation}\label{eq:condition}
\lim_{N\rightarrow +\infty} \varphi^N_{\bm{\mathcal{F}}_c^{\mathrm{R}}}\big(\Lambda^{[m]},\bm{\xi}\big)=\lim_{N\rightarrow +\infty}\ln \bm{\mathcal{N}}_1\Big(\mathrm{Prj}_{\Lambda^{[m]}}^{\bm{\mathcal{F}}_c^{\mathrm{R}}},\frac{\bm{\xi}_{\Lambda^{[m]}}}{8},2N\Big)=0,
\end{equation}
it then holds that
\begin{equation}\label{eq:main2}
\lim_{N\rightarrow +\infty}   \mathrm{Pr}\Big\{\sup_{{\bf f} \in\bm{\mathcal{F}}_c^\mathrm{R}}
\big|{\bf E}{\bf f}-{\bf E}_N{\bf f}\big|\not\leq {\bm \xi} \Big\}=0.
\end{equation}
\end{theorem}

In this theorem, we obtain an upper bound of the joint probability of the event that $\sup_{{\bf f} \in\bm{\mathcal{F}}_c^\mathrm{R}}
\big|{\bf E}{\bf f}-{\bf E}_N{\bf f}\big|\not\leq {\bm \xi}$ and show that the consistency of each task in MTL can be guaranteed if condition \eqref{eq:condition} is valid. We are concerned with two aspects of the theorem:
\begin{itemize}
\item the RHS of \eqref{eq:main} implies that given $N<\infty$, a smaller value of EDDM $\varphi^N_{\bm{\mathcal{F}}_c^{\mathrm{R}}}(\Lambda^{[m]},\bm{\xi})$ will lead to a sharper bound, which is in accordance with the argument that the negative EDDM means that the task groups benefit from each other (see Section \ref{sec:measure}).
\item The asymptotic convergence of the generalization bound is determined by two factors: 1) EDDM $\varphi^N_{\bm{\mathcal{F}}_c^{\mathrm{R}}}\big(\Lambda^{[m]},\bm{\xi}\big)$; and 2) CPUEN $\ln \bm{\mathcal{N}}_1\big(\mathrm{Prj}_{\Lambda^{[m]}}
    ^{\bm{\mathcal{F}}_c^{\mathrm{R}}},\bm{\xi}_{\Lambda^{[m]}}/8,2N\big)$. In particular, according to the classical results of STL (see Theorem 2.3 \& Definition 2.5 of \citep{Mendelson03}), if UEN for each task $\mathcal{Z}^{[m]}$ satisfies that $\frac{\ln\mathcal{N}_1(\mathcal{F}^{[m]},\xi^{[m]}/8,2N)}{N}$ converges to {\it zero} when $N$ goes to {\it infinity}, the second equality of \eqref{eq:condition} holds. Note that the convergence of $\varphi^N_{\bm{\mathcal{F}}_c^{\mathrm{R}}}\big(\Lambda^{[m]},\bm{\xi}\big)$ is determined by the inherent characteristics of MRTs, {\it e.g.,} distributions of tasks, selection of function classes, and regularization terms.

\end{itemize}

\begin{remark}\label{rem:examine}
Moreover, these theoretical findings cause us to preliminarily examine whether the combination of tasks, function classes, and regularization terms is suitable for the ERM-based RMTL according to the rules that
$$\Upsilon(\bm{\Lambda})=\sum\limits_{m=1}^M\sum\limits_{\Lambda^{[m]}\subset\bm{\Lambda}} \phi_{\bm{\mathcal{F}}}(\Lambda^{[m]},\bm{\xi})< \frac{1}{2},$$ and $$\lim\limits_{N\rightarrow +\infty} \varphi^N_{\bm{\mathcal{F}}_c^{\mathrm{R}}}\big(\Lambda^{[m]},\bm{\xi}\big)=0$$ with $\varphi^N_{\bm{\mathcal{F}}_c^{\mathrm{R}}}(\Lambda^{[m]},\bm{\xi})\leq 0$ for any $\Lambda^{[m]}\subset\bm{\Lambda}$ ($1\leq m\leq M$).

\end{remark}

%---------------------------------------------------

\section{Generalization Bounds with Covariance Information}\label{sup:covariance}

As discussed in Section \ref{sec:measure}, since ODDM detects the dependence between two task groups, the bound \eqref{eq:bound.A} cannot reflect how the individual relatedness between two tasks affects the generalization performance of RMTL for more than two tasks. Here, we consider the generalization results based on the covariance information between every two tasks.

%\begin{equation}\label{eq:sym.condition.1.A}
 % \sum_{m=1}^M\sum_{\Lambda^{[m]}\subset\bm{\Lambda}}\phi_{\bm{\mathcal{F}}}(\Lambda^{[m]},\bm{\xi})+
%\frac{4m(b-a)^2}{N\sum\limits_{i\in \Lambda^{[m]}}(\xi^{[i]})^2}\leq \frac{1}{2},
%\end{equation}

\begin{theorem}\label{thm:bound}
Follow the notations of Theorem \ref{thm:bound.A}. Given $\bm{\xi}=(\xi^{[1]},\cdots,\xi^{[M]})^T>{\bf 0}$ and for any $N\in\mathbb{N}$ such that
\begin{equation}\label{eq:sym.condition.1}
 N\geq \frac{8\Gamma_2}{1-2(\Upsilon(\bm{\Lambda})+\Upsilon_2)},
\end{equation}
then there holds that
\begin{align}\label{eq:bound}
  \mathrm{Pr}\left\{\sup_{{\bf f} \in\bm{\mathcal{F}}_c^\mathrm{R}}
\big|{\bf E}{\bf f}-{\bf E}_N{\bf f}\big|> {\bm \xi} \right\}\leq 2^{M+2} \bm{\mathcal{N}}_1\big(\bm{\mathcal{F}}_c^{\mathrm{R}},\bm{\xi}/8,2N\big) \exp\left\{\frac{-N\sum_{m=1}^M(\xi^{[m]})^2}{32M^2(b-a)^2}\right\},
\end{align}
where
\begin{equation}\label{eq:Gamma}
  \Gamma_2:=\sum_{m=1}^M\sum_{\Lambda^{[m]}\subset\bm{\Lambda}}
\frac{m(b-a)^2}{\Big(\sum\limits_{i\in \Lambda^{[m]}}\xi^{[i]}\Big)^2},
\end{equation}
and
\begin{equation}\label{eq:Upsilon}
\Upsilon_2:=\sum_{m=1}^M\sum_{\Lambda^{[m]}\subset\bm{\Lambda}}
 \frac{8\mathop{\sum\limits_{i_1<i_2}}
 \limits_{i_1,i_2\in\Lambda^{[m]}}
\mathrm{Cov}_{\bm{\mathcal{F}}}
  \left(i_1,i_2\right)}{\Big(\sum\limits_{i\in \Lambda^{[m]}}\xi^{[i]}\Big)^2}.
\end{equation}
%and
%\begin{equation}\label{eq:Upsilon}
% \Upsilon_2:=\sum_{m=1}^M\sum_{\Lambda^{[m]}\subset\bm{\Lambda}}
% \frac{8\mathop{\sum\limits_{i_1<i_2}}
% \limits_{i_1,i_2\in\Lambda^{[m]}}
%\mathrm{Cov}_{\bm{\mathcal{F}}}
%  \left(i_1,i_2\right)}{\Big(\sum\limits_{i\in \Lambda^{[m]}}\xi^{[i]}\Big)^2}.
%\end{equation}
\end{theorem}
Compared to Theorem \ref{thm:bound.A}, the condition \eqref{eq:sym.condition.1} incorporates the quantity $\Upsilon_2$ which is related to the covariance information. Actually, the quantity $\Upsilon_2$ is derived by replacing $\sum\limits_{i\in \Lambda^{[m]}}(\xi^{[i]})^2$ with $\big(\sum\limits_{i\in \Lambda^{[m]}}\xi^{[i]}\big)^2$ as shown in the proof of Lemma \ref{lem:Chebyshev}. From the condition \eqref{eq:sym.condition.1}, we can find that the bound \eqref{eq:bound} is valid when $\Upsilon(\bm{\Lambda})+\Upsilon_2<1/2$, which means that if the synergetic effect is the main group relatedness in the learning process and some of the correlations between tasks are negative, the learning process will perform well with a small sample size $N$. \citet{Zhang10convex} have highlighted the necessity of the negative correlation and
pointed out that the negative correlation is helpful to reduce the search space in MTL, which is in accordance with our theoretical findings.

However, when $M=1$, the bound \eqref{eq:bound} coincides with the canonical results in STL if and only if the quantity $\mathrm{Cov}_{\bm{\mathcal{F}}}
  \left(i,i\right)$ equals to {\it zero}, {\it i.e.,} the random variable ${\bf z}$ of the task $\mathcal{Z}^{[i]}$ takes a constant with the probability of {\it one}. Since this setting is far away from the practical scenario, unlike the result \eqref{eq:main}, the bound \eqref{eq:bound} that encodes covariance information cannot reflect the transition from STL to MTL.

%---------------------------------------------------

\section{Conclusion}

In this paper, we apply the vector-valued framework to study the generalization performance of RMTL and analyze the relationship between the task-group relatedness and the properties of RMTL.
In particular, we introduce two types of task-group relatedness: ODDM and EDDM, and we present CPUEN to measure the complexity of the regularized vector-valued function class
$\bm{\mathcal{F}}_c^{\mathrm{R}}$. By applying the specific deviation and symmetrization inequalities to the vector-valued framework, we obtain the generalization bound for RMTL and provide a sufficient condition to guarantee the consistency of each task in the simultaneous learning process of MRTs. Finally, we show that the theoretical findings of this paper can examine whether the task settings are suitable for the RMTL mechanism

Based on the theoretical findings, we summarize the relationship between the generalization properties of RMTL and the task-group relatedness as follows:

\begin{itemize}
%\item Theoretically, the use of ODDM and EDDM is enough to support the analysis of the generalization properties of MTL.
%
%\item The negative ODDM and EDDM imply that the two groups of tasks will benefit from each other, {\it i.e.,} the relatedness is of synergy effect.
%
\item ODDM is related to the sample size and validity of RMTL (see Theorem \ref{thm:bound.A}). We first prove that the condition of $\Upsilon(\bm{\Lambda})<\frac{1}{2}$ is necessary for the validity of RMTL and then show that if almost any pair of task groups $\Lambda^{[m]}$ and $\overline{\Lambda^{[m]}}$ predominantly mutually promote, the required sample size $N$ of each task in RMTL will be smaller than that of STL for each task. The sample size will also not increase dramatically, regardless of a large number of MRTs
    (see Remarks \ref{rem:ODDM1} \& \ref{rem:ODDM2}).\vspace{-1mm}

\item EDDM affects the generalization performance of RMTL as follows: 1) a negative EDDM provides a sharper generalization bound; and 2) the asymptotic behavior of EDDM also affects the consistency of the task (see Theorem \ref{thm:main}). \vspace{-1mm}
\item The existence of a negative correlation between two tasks is necessary for MTL, which is in accordance with the relevant argument of \citep{Zhang10convex}.
%\item EDDM affects the generalization performance of RMTL: 1) the negative EDDM with $N<\infty$ provides a sharper generalization bound; 2) the condition that EDDM degenerates to {\it zero} when the sample size $N$ goes to {\it infinity} is
%%
%\item MTL will perform well with a smaller size of samples if most of task groups are of the synergic relatedness with others.
%
\end{itemize}
In summary, synergistic task-group relatedness is beneficial to the generalization performance of RMTL. In future works, we will focus on the practical applications of the theoretical findings, for instance by improving the empirical computations of ODDM and EDDM (see Remark \ref{rem:examine}) and designing the regularization term for RMTL based on the task-group relatedness.

\bibliographystyle{plainnat}
\bibliography{Ref-Multitask}
\newpage
%----------------------------------------------------------------

\appendix

\section{Deviation Inequalities for Random Vectors }\label{sup:deviation}

To obtain the generalization bounds for RMTL, we need to consider the deviation inequalities for random vectors. The following lemma is derived from \citep{chen2013concentration}.

Let ${\bf s}_n=(s^{[1]}_n,\cdots,s^{[M]}_n)^T\in\mathbb{R}^M$ ($1\leq n\leq N$) be $N$ i.i.d. random vectors such that
\begin{equation}\label{eq:dev.cond1}
  \sum_{m=1}^M s_n^{[m]}\leq 1, \quad \mbox{for $n=1,\cdots,N$,}
\end{equation}
and
\begin{equation}\label{eq:dev.cond2}
 s_n^{[m]}\geq 0, \quad \mbox{for $1 \leq n\leq N$ and $1\leq m\leq M$. }
\end{equation}
Note that the components $s_n^{[1]},\cdots,s_n^{[M]}$ of ${\bf s}_n$ are not necessarily independent. The mean $\bm{\mu}=(\mu^{[1]},\cdots,\mu^{[M]})^T$ of random vectors ${\bf s}_n$ is expressed as
\begin{equation}\label{eq:mean}
 \mu^{[m]}=\mathrm{E}^{[m]}s_n^{[m]},\quad \mbox{for $1\leq m\leq M$. }
\end{equation}

\begin{lemma}\label{lem:dev}
For any $\bm{\xi}=(\xi^{[1]},\cdots,\xi^{[M]})^T>0$ such that $\sum_{m=1}^M(\mu^{[m]}+\xi^{[m]})<1$, then there holds that
\begin{equation}\label{eq:lem.dev}
  \mathrm{Pr}\left\{\Big|\frac{1}{N}\sum_{n=1}^N{\bf s}_n - \bm{\mu}\Big| > \bm{\xi} \right\}\leq 2^M \exp\left\{-2N\sum_{m=1}^M(\xi^{[m]})^2\right\}.
\end{equation}
\end{lemma}
Moreover, since the vector-valued function ${\bf f}$ has the range $[a,b]$, let
\begin{equation}\label{eq:dev.pr1}
  s_n^{[m]}:=\frac{f^{[m]}({\bf z}_n^{[m]})-a}{M(b-a)},\qquad 1\leq n\leq N,\;1\leq m\leq M,
\end{equation}
and then
\begin{equation}\label{eq:dev.pr2}
  \mathrm{Pr}\Big\{\big|{\bf E}_N{\bf f}-{\bf E}{\bf f}\big|>\bm{\xi}\Big\}
  =\mathrm{Pr}\left\{\Big|\frac{1}{N}\sum_{n=1}^N {\bf s}_n-\frac{{\bf E}{\bf f}-{\bf a}}{M(b-a)}\Big|>\frac{\bm{\xi}}{M(b-a)}\right\},
\end{equation}
where ${\bf a}= (a,\cdots,a)^T\in\mathbb{R}^M$. Thus, the combination of Lemma \ref{lem:dev} and \eqref{eq:dev.pr2} leads to a Hoeffding-type deviation inequality for random vectors.
\begin{theorem}\label{thm:dev}
Given a bounded vector-valued function ${\bf f}=(f^{[1]},\cdots,f^{[M]})^T$ with the range $[a,b]$, there holds that for any $\bm{\xi}=(\xi^{[1]},\cdots,\xi^{[M]})>0$,
\begin{equation}\label{eq:dev}
  \mathrm{Pr}\left\{\big|{\bf E}_N{\bf f} - {\bf E}{\bf f}\big| > \bm{\xi} \right\}\leq 2^M \exp\left\{-2N\sum_{m=1}^M\frac{(\xi^{[m]})^2}{M^2(b-a)^2}\right\}.
\end{equation}

\end{theorem}

%%%%%%%%%%%%%%

\section{Symmetrization Inequalities for Random Vectors}\label{sup:symmetrization}

\subsection{Chebyshev Inequalities for Random Vectors}

\begin{definition}\label{def:psi}
Assume that $\mathcal{Z}^{[1]},\cdots,\mathcal{Z}^{[M]}$ are $M$ distributions on $\mathbb{R}$.
Let $\bm{\Lambda}:=\{1,\cdots,M\}$ be an index set and $\Lambda^{[m]}$ be a subset of $\bm{\Lambda}$ with the cardinality of $m$. For any $\Lambda^{[m]}\subset\bm{\Lambda}$ and any $\bm{\xi}=(\xi^{[1]},\cdots,\xi^{[M]})^T>{\bf 0}$, define
\begin{align}\label{eq:psi}
\psi(\Lambda^{[m]},\bm{\xi}):=
&\mathrm{Pr}\Big\{   \{s^{[i]}>\xi^{[i]}\}_{i\in\Lambda^{[m]}} \big|
\{s^{[i]}\leq\xi^{[i]}\}_{i\in\overline{\Lambda^{[m]}}}   \Big\}-\mathrm{Pr}\Big\{   \{s^{[i]}>\xi^{[i]}\}_{i\in\Lambda^{[m]}}  \Big\}.
\end{align}
where $s^{[i]}$ is the non-negative random variable of the task $\mathcal{Z}^{[i]}$, and $\overline{\Lambda^{[m]}}$ stands for the complementary set of $\Lambda^{[m]}$ with $\Lambda^{[m]}\cup\overline{\Lambda^{[m]}}=\bm{\Lambda}$.
\end{definition}

\begin{lemma}\label{lem:Chebyshev.A}
Let ${\bf s}=(s^{[1]},\cdots,s^{[M]})^T$ be a random vector with nonnegative elements and $\bm{\Lambda}=\{1,\cdots,M\}$ be an index set.
For any $\bm{\xi}=(\xi^{[1]},\cdots,\xi^{[M]})^T>0$, then there holds that
\begin{equation}\label{eq:Chebyshev.A}
  \mathrm{Pr}\left\{   {\bf s}\not\leq \bm{\xi} \right\}\leq \sum_{m=1}^M\sum_{\Lambda^{[m]}\subset\bm{\Lambda}}\left(\psi(\Lambda^{[m]},\bm{\xi})+
  \frac{\sum\limits_{i\in \Lambda^{[m]}}\mathrm{E}^{[i]}\big\{(s^{[i]})^2\big\}}{\sum\limits_{i\in \Lambda^{[m]}}(\xi^{[i]})^2}\right),
\end{equation}
where ${\bf s}\not\leq \bm{\xi}$ means that there is at least one index $m\in\bm{\Lambda}$ such that $s^{[m]}>\xi^{[m]}$, and $\Lambda^{[m]}$ stands for an index set with the cardinality of $m$.

\end{lemma}

\begin{lemma}\label{lem:Chebyshev}
Let ${\bf s}=(s^{[1]},\cdots,s^{[M]})^T$ be a random vector with nonnegative elements and $\bm{\Lambda}=\{1,\cdots,M\}$ be an index set.
For any $\bm{\xi}=(\xi^{[1]},\cdots,\xi^{[M]})^T>0$, then there holds that
\begin{equation}\label{eq:Chebyshev}
  \mathrm{Pr}\left\{   {\bf s}\not\leq \bm{\xi} \right\}\leq \sum_{m=1}^M\sum_{\Lambda^{[m]}\subset\bm{\Lambda}}\left(\psi(\Lambda^{[m]},\bm{\xi})+
  \frac{\sum\limits_{i\in \Lambda^{[m]}}\mathrm{E}^{[i]}\big\{(s^{[i]})^2\big\}+2\mathop{\sum\limits_{i<j}}\limits_{i,j\in\Lambda^{[m]}}
  \mathrm{E}\big\{s^{[i]}s^{[j]}\big\}}{\Big(\sum\limits_{i\in \Lambda^{[m]}}\xi^{[i]}\Big)^2}\right),
\end{equation}
where ${\bf s}\not\leq \bm{\xi}$ means that there is at least one index $m\in\bm{\Lambda}$ such that $s^{[m]}>\xi^{[m]}$, and $\Lambda^{[m]}$ stands for an index set with the cardinality of $m$.

\end{lemma}

\subsection{Symmetrization Inequalities}

By applying ODDM, we can develop the symmetrization inequality for MTL as follows:

\begin{theorem}\label{thm:sym.A}
Assume that $\bm{\mathcal{F}}$ is a vector-valued function class with the range $[a,b]$.
For any $\bm{\xi}\geq {\bf 0}$ such that
\begin{equation}\label{eq:sym.condition.A}
   N\geq \frac{8\Gamma(\bm{\Lambda})}{1-2\Upsilon(\bm{\Lambda})},
\end{equation}
then there holds that
\begin{equation}\label{eq:sym.A}
  \mathrm{Pr}\left\{\sup_{{\bf f} \in\bm{\mathcal{F}}}
\big|{\bf E}{\bf f}-{\bf E}_N{\bf f}\big|> {\bm \xi} \right\}\leq
2  \mathrm{Pr}\left\{\sup_{{\bf f} \in\bm{\mathcal{F}}}
\big|{\bf E}'_N{\bf f}-{\bf E}_N{\bf f}\big|> \frac{{\bm \xi}}{2} \right\},
\end{equation}
where
\begin{equation*}%\label{eq:Gamma0}
  \Gamma(\bm{\Lambda}):=\sum_{m=1}^M\sum_{\Lambda^{[m]}\subset\bm{\Lambda}}\frac{m(b-a)^2}{\sum\limits_{i\in \Lambda^{[m]}}(\xi^{[i]})^2},\;\;\Upsilon(\bm{\Lambda}):=\sum_{m=1}^M\sum_{\Lambda^{[m]}\subset\bm{\Lambda}}
  \phi_{\bm{\mathcal{F}}}(\Lambda^{[m]},\bm{\xi}),
\end{equation*}
$\bm{\Lambda}=\{1,\cdots,M\}$ is an index set and $\Lambda^{[m]}$ is a subset of $\bm{\Lambda}$ with the cardinality of $m$.
\end{theorem}

The following is the symmetrization result incorporating the covariance information between every two tasks.

\begin{theorem}\label{thm:sym}
Assume that $\mathcal{F}$ is a vector-valued function class with the range $[a,b]$.
For any $\bm{\xi}=(\xi^{[1]},\cdots,\xi^{[M]})^T>0$ such that
\begin{equation}\label{eq:sym.condition}
  N\geq \frac{8\Gamma_2}{1-2(\Upsilon(\bm{\Lambda})+\Upsilon_2)},
\end{equation}
then there holds that
\begin{equation}\label{eq:sym}
  \mathrm{Pr}\left\{\sup_{{\bf f} \in\bm{\mathcal{F}}}
\big\{|{\bf E}{\bf f}-{\bf E}_N{\bf f}\big\}|> {\bm \xi} \right\}\leq
2  \mathrm{Pr}\left\{\sup_{{\bf f} \in\bm{\mathcal{F}}}
\big\{|{\bf E}'_N{\bf f}-{\bf E}_N{\bf f}|\big\}> \frac{{\bm \xi}}{2} \right\},
\end{equation}
where
\begin{equation*}%\label{eq:Gamma}
  \Gamma_2:=\sum_{m=1}^M\sum_{\Lambda^{[m]}\subset\bm{\Lambda}}
\frac{m(b-a)^2}{\Big(\sum\limits_{i\in \Lambda^{[m]}}\xi^{[i]}\Big)^2},\;\;\Upsilon_2:=\sum_{m=1}^M\sum_{\Lambda^{[m]}\subset\bm{\Lambda}}
 \frac{8\mathop{\sum\limits_{i_1<i_2}}
 \limits_{i_1,i_2\in\Lambda^{[m]}}
\mathrm{Cov}_{\bm{\mathcal{F}}}
  \left(i_1,i_2\right)}{\Big(\sum\limits_{i\in \Lambda^{[m]}}\xi^{[i]}\Big)^2}
\end{equation*}
and
$\mathrm{Cov}_{\bm{\mathcal{F}}}
  \left(i_1,i_2\right)$ is defined as
  \begin{equation}\label{eq:cov}
 \mathrm{Cov}_{\bm{\mathcal{F}}}(i,j) := \max_{(f^{[1]},\cdots,f^{[M]})^T\in\bm{\mathcal{F}}} \mathrm{Cov}
  \left(f^{[i]}({\bf z}^{[i]}),f^{[j]}({\bf z}^{[j]})\right)
\end{equation}
with ${\bf z}^{[i]}$ and ${\bf z}^{[j]}$ ($1\leq i,j\leq M$) being the random variables of the tasks $\mathcal{Z}^{[i]}$ and $\mathcal{Z}^{[j]}$, respectively.

\end{theorem}

%%%

%%%%%

\section{Proofs of Main Results}\label{sup:proofs}

\subsection{Proof of Lemma \ref{lem:dev}}

{\it Proof of Lemma \ref{lem:dev}.} Let ${\bf t}= \big|\frac{1}{N}\sum_{n=1}^N{\bf s}_n - \bm{\mu}\big|$. The event $|{\bf t}|> \bm{\xi}$ contains $2^M$ possibilities: for any $1\leq m\leq M$, there are $m$ components of the vector ${\bf t}$ such that $t^{[i_k]}>\xi^{[i_k]}$ ($1\leq k\leq m$) and the rest are of the case that $t^{[i_k]}<-\xi^{[i_k]}$ ($1\leq k\leq M-m$). For convenience, we also denote $\{\mathcal{P}_i\}_{i=1}^{2^M}$ as the collection of all $2^M$ possibilities.

According to Theorem 1 in \citep{chen2013concentration}, the following result is valid for any possibility $\mathcal{P}_i$ ($1\leq i\leq 2^M$):
\begin{equation}\label{eq:lem.dev.pr1}
  \mathrm{Pr}\left\{\mathcal{P}_i \right\}
  \leq \prod_{m=0}^M\left(\frac{\mu^{[m]}}{p^{[m]}}\right)^{p^{[m]}N},
\end{equation}
where $p^{[m]}=\mu^{[m]}+\xi^{[m]}$ ($m=1,\cdots,M$), $\mu_0=1-\sum_{m=1}^M\mu^{[m]}$ and $p_0=1-\sum_{m=1}^Mp^{[m]}$.
Then, we have
\begin{equation}\label{eq:lem.dev.1}
  \mathrm{Pr}\left\{\Big|\frac{1}{N}\sum_{n=1}^N{\bf s}_n - \bm{\mu}\Big| > \bm{\xi} \right\}\leq 2^M \prod_{m=0}^M\left(\frac{\mu^{[m]}}{p^{[m]}}\right)^{p^{[m]}N}.
\end{equation}

Then, consider
\begin{align}\label{eq:lem.dev.pr2}
  \prod_{m=0}^M\left(\frac{\mu^{[m]}}{p^{[m]}}\right)^{p^{[m]}N}=&\exp\left\{N\sum_{m=0}^Mp^{[m]}
                                                       \log\Big(\frac{\mu^{[m]}}{p^{[m]}}\Big)\right\}\nonumber\\
=&\exp\left\{N\left(\Big(1-\sum_{m=1}^Mp^{[m]}\Big)\log\Big(\frac{1-\sum_{m=1}^M\mu^{[m]}}
{1-\sum_{m=1}^Mp^{[m]}}\Big)+\sum_{m=1}^Mp^{[m]}
\log\Big(\frac{\mu^{[m]}}{p^{[m]}}\Big)\right)\right\}\nonumber\\
\leq&\exp\left\{N\left(\sum_{m=1}^M\Big(1-p^{[m]}\Big)\log\Big(\frac{1-\mu^{[m]}}
{1-p^{[m]}}\Big)+\sum_{m=1}^Mp^{[m]}
\log\Big(\frac{\mu^{[m]}}{p^{[m]}}\Big)\right)\right\}\quad (*)\nonumber\\
=&\exp\left\{-N\sum_{m=1}^M \int_{\mu^{[m]}}^{p^{[m]}}\left(\frac{p^{[m]}}{x}-\frac{1-p^{[m]}}{1-x}\right)dx\right\}\nonumber\\
=&\exp\left\{-N\sum_{m=1}^M \int_{\mu^{[m]}}^{p^{[m]}}\frac{p^{[m]}-x}{x(1-x)}dx\right\}\nonumber\\
\leq&\exp\left\{-N\sum_{m=1}^M 4\int_{\mu^{[m]}}^{p^{[m]}}(p^{[m]}-x)dx\right\}\nonumber\\
=&\exp\left\{-2N\sum_{m=1}^M(p^{[m]}-\mu^{[m]})^2\right\}=\exp\left\{-2N\sum_{m=1}^M(\xi^{[m]})^2\right\},
  \end{align}
because $x(1-x)\leq 1/4$ for any $x\in\mathbb{R}$, and the step $(*)$ is followed from the fact that the function $f$ is subadditive if $f$ is concave and $f(0)\geq 0$. \hfill$\blacksquare$

%%%%

\subsection{Proof of Lemma \ref{lem:Chebyshev.A}}

{\it Proof of Lemma \ref{lem:Chebyshev.A}.} Given $M$ tasks $\mathcal{Z}^{(1)},\cdots,\mathcal{Z}^{[M]}$ and a vector-valued function class $\bm{\mathcal{F}}$,
let $\bm{\Lambda}:=\{1,\cdots,M\}$ be an index set and $\Lambda^{[m]}$ be a subset of $\bm{\Lambda}$ with the cardinality of $m$. For any $\Lambda^{[m]}\subset\bm{\Lambda}$ and any $\bm{\xi}=(\xi^{(1)},\cdots,\xi^{[M]})^T>{\bf 0}$, define
\begin{align}\label{eq:phi}
\psi(\Lambda^{[m]},\bm{\xi}):=
&\mathrm{Pr}\big\{   \{s^{[i]}>\xi^{[i]}\}_{i\in\Lambda^{[m]}} \big|
\{s^{[i]}\leq\xi^{[i]}\}_{i\in\overline{\Lambda^{[m]}}}   \big\}-\mathrm{Pr}\big\{   \{s^{[i]}>\xi^{[i]}\}_{i\in\Lambda^{[m]}}  \big\}.
\end{align}
Then, the event ${\bf s}\not\leq \bm{\xi}$ contains the following possibilities:
\begin{itemize}
  \item $\mathcal{P}^{[1]}$: there is only one index $\{i\}= \Lambda^{[1]}\subset\bm{\Lambda}$ satisfying that $s^{[i]}>\xi^{[i]}$;
  \item $\mathcal{P}^{[m]}$: there are only $m$ ($1<m<M$) indices $\{i^{[1]},\cdots,i^{[m]}\}=\Lambda^{[m]}\subset\bm{\Lambda}$ satisfying that $s^{[i_k]}>\xi^{[i_k]}$ ($1\leq k\leq m$);
  \item $\mathcal{P}^{[M]}$: $s^{[m]}>\xi^{[m]}$ holds for any $1\leq m\leq M$.
\end{itemize}
Thus, we have
\begin{equation}\label{eq:Chebyshev.pr1.A}
  \mathrm{Pr}\left\{   {\bf s}\not\leq \bm{\xi} \right\}= \mathrm{Pr}\{\mathcal{P}^{[1]}\}+\cdots+\mathrm{Pr}\{\mathcal{P}^{[M]}\}.
\end{equation}
According to Chebyshev's inequality and \eqref{eq:phi}, we have
\begin{equation}\label{eq:Chebyshev.pr2.A}
  \mathrm{Pr}\{\mathcal{P}^{[1]}\}= \sum_{m=1}^M\left(\psi(\{m\},\bm{\xi})+\mathrm{Pr}\{s^{[m]}>\xi^{[m]}\}\right)\leq \sum_{m=1}^M \left(\psi(\{m\},\bm{\xi})+\frac{\mathrm{E}\{(s^{[m]})^2\}}{(\xi^{[m]})^2}\right),
\end{equation}
 and for any $2\leq m\leq M$,
 \begin{align}\label{eq:Chebyshev.pr3.A}
  \mathrm{Pr}\{\mathcal{P}^{[m]}\}=& \sum_{\Lambda^{[m]}\subset\bm{\Lambda}}\left(\psi(\Lambda^{[m]},\bm{\xi})+ \mathrm{Pr}\{s^{[i]}>\xi^{[i]}:i\in\Lambda^{[m]}\}\right)\nonumber\\
  =& \sum_{\Lambda^{[m]}\subset\bm{\Lambda}}\left(\psi(\Lambda^{[m]},\bm{\xi})+ \mathrm{Pr}\{(s^{[i]})^2>(\xi^{[i]})^2\;:\;i\in\Lambda^{[m]}\}\right)\nonumber\\
  \leq& \sum_{\Lambda^{[m]}\subset\bm{\Lambda}}\left(\psi(\Lambda^{[m]},\bm{\xi})+ \mathrm{Pr}\left\{\sqrt{\sum_{i\in\Lambda^{[m]}}(s^{[i]})^2}
  >\sqrt{\sum_{i\in\Lambda^{[m]}}(\xi^{[i]})^2}
  \right\}\right)\nonumber\\
  \leq & \sum_{\Lambda^{[m]}\subset\bm{\Lambda}}\left(\psi(\Lambda^{[m]},\bm{\xi})+ \frac{\mathrm{E}\left\{\sum\limits_{i\in\Lambda^{[m]}}(s^{[i]})^2\right\}}
  {\sum\limits_{i\in\Lambda^{[m]}}(\xi^{[i]})^2}\right)\nonumber\\
   = & \sum_{\Lambda^{[m]}\subset\bm{\Lambda}}\left(\psi(\Lambda^{[m]},\bm{\xi})+ \frac{\sum\limits_{i\in\Lambda^{[m]}}\mathrm{E}\{s^{[i]}\}^2
  }
  {\sum\limits_{i\in\Lambda^{[m]}}(\xi^{[i]})^2}\right).
\end{align}
The combination of \eqref{eq:Chebyshev.pr1.A}, \eqref{eq:Chebyshev.pr2.A} and \eqref{eq:Chebyshev.pr3.A} leads to the result \eqref{eq:Chebyshev.A}. This completes the proof.
\hfill$\blacksquare$

\subsection{Proof of Lemma \ref{lem:Chebyshev}}

{\it Proof of Lemma \ref{lem:Chebyshev}.} The event ${\bf s}\not\leq \bm{\xi}$ contains the following possibilities:
\begin{itemize}
  \item $\mathcal{P}^{[1]}$: there is only one index $\{i\}= \Lambda^{[1]}\subset\bm{\Lambda}$ satisfying that $s^{[i]}>\xi^{[i]}$;
  \item $\mathcal{P}^{[m]}$: there are only $m$ ($1<m<M$) indices $\{i^{[1]},\cdots,i^{[m]}\}=\Lambda^{[m]}\subset\bm{\Lambda}$ satisfying that $s^{[i_k]}>\xi^{[i_k]}$ ($1\leq k\leq m$);
  \item $\mathcal{P}^{[M]}$: $s^{[m]}>\xi^{[m]}$ holds for any $1\leq m\leq M$.
\end{itemize}
Thus, we have
\begin{equation}\label{eq:Chebyshev.pr1}
  \mathrm{Pr}\left\{   {\bf s}\not\leq \bm{\xi} \right\}= \mathrm{Pr}\{\mathcal{P}^{[1]}\}+\cdots+\mathrm{Pr}\{\mathcal{P}^{[M]}\}.
\end{equation}
According to Chebyshev's inequality and \eqref{eq:phi}, we have
\begin{equation}\label{eq:Chebyshev.pr2}
  \mathrm{Pr}\{\mathcal{P}^{[1]}\}= \sum_{m=1}^M\left(\psi(\{m\},\bm{\xi})+\mathrm{Pr}\{s^{[m]}>\xi^{[m]}\}\right)\leq \sum_{m=1}^M \left(\psi(\{m\},\bm{\xi})+\frac{\mathrm{E}\{(s^{[m]})^2\}}{(\xi^{[m]})^2}\right),
\end{equation}
 and for any $2\leq m\leq M$,
 \begin{align}\label{eq:Chebyshev.pr3}
  \mathrm{Pr}\{\mathcal{P}^{[m]}\}=& \sum_{\Lambda^{[m]}\subset\bm{\Lambda}}\left(\psi(\Lambda^{[m]},\bm{\xi})+ \mathrm{Pr}\{s^{[i]}>\xi^{[i]}:i\in\Lambda^{[m]}\}\right)\nonumber\\
  \leq& \sum_{\Lambda^{[m]}\subset\bm{\Lambda}}\left(\psi(\Lambda^{[m]},\bm{\xi})+ \mathrm{Pr}\left\{\sum_{i\in\Lambda^{[m]}}(s^{[i]})
  >\sum_{i\in\Lambda^{[m]}}(\xi^{[i]})
  \right\}\right)\nonumber\\
  \leq & \sum_{\Lambda^{[m]}\subset\bm{\Lambda}}\left(\psi(\Lambda^{[m]},\bm{\xi})+ \frac{\mathrm{E}\left\{\Big(\sum\limits_{i\in\Lambda^{[m]}}s^{[i]}\Big)^2\right\}}
  {\Big(\sum\limits_{i\in\Lambda^{[m]}}\xi^{[i]}\Big)^2}\right)\nonumber\\
   = & \sum_{\Lambda^{[m]}\subset\bm{\Lambda}}\left(\psi(\Lambda^{[m]},\bm{\xi})+ \frac{\sum\limits_{i\in\Lambda^{[m]}}\mathrm{E}\{(s^{[i]})^2\}
  +2\mathop{\sum\limits_{i,j\in\Lambda^{[m]}}}\limits_{i<j}\mathrm{E}\{s^{[i]}s^{[j]}\}}
  {\Big(\sum\limits_{i\in\Lambda^{[m]}}\xi^{[i]}\Big)^2}\right).
\end{align}
The combination of \eqref{eq:Chebyshev.pr1}, \eqref{eq:Chebyshev.pr2} and \eqref{eq:Chebyshev.pr3} leads to the result \eqref{eq:Chebyshev}. This completes the proof.
\hfill$\blacksquare$

\subsection{Proof of Theorem \ref{thm:sym.A}}

{\it Proof of Theorem \ref{thm:sym.A}.} Let ${\bf f}_N=(\widehat{f}^{[1]},\cdots,\widehat{f}^{[M]})^T$ be the vector-valued function achieving the supremum
\begin{equation*}%\label{eq:sym.pr1.A}
  \sup_{{\bf f} \in\bm{\mathcal{F}}}
\big|{\bf E}{\bf f}-{\bf E}_N{\bf f}\big|.
\end{equation*}
According to the triangle inequality, we have
\begin{align}\label{eq:sym.pr0.A}
  |{\bf E}{\bf f}_N-{\bf E}_N{\bf f}_N|-|{\bf E}'_N{\bf f}_N-{\bf E}{\bf f}_N|\leq |{\bf E}'_N{\bf f}_N-{\bf E}_N{\bf f}_N|,
\end{align}
and thus
\begin{align}\label{eq:sym.pr1.A}
  {\bf 1}_{\left\{|{\bf E}{\bf f}_N-{\bf E}_N{\bf f}_N|> \bm{\xi}\right\}}
  {\bf 1}_{\left\{|{\bf E}{\bf f}_N-{\bf E}'_N{\bf f}_N|\leq \bm{\xi}/2\right\}}
  =&{\bf 1}_{\left\{|{\bf E}{\bf f}_N-{\bf E}_N{\bf f}_N|> \bm{\xi}\right\}
  \wedge \left\{|{\bf E}'_N{\bf f}_N-{\bf E}{\bf f}_N|\leq \bm{\xi}/2\right\} }\nonumber\\
  \leq & {\bf 1}_{\left\{|{\bf E}'_N{\bf f}_N-{\bf E}_N{\bf f}_N|> \bm{\xi}/2\right\}}.
\end{align}
Taking expectations with respect to the ghost samples gives
\begin{align}\label{eq:sym.pr2.A}
  {\bf 1}_{\left\{|{\bf E}{\bf f}_N-{\bf E}_N{\bf f}_N|> \bm{\xi}\right\}}
  \mathrm{Pr}'\left\{\big|{\bf E}{\bf f}_N-{\bf E}'_N{\bf f}_N\big|\leq \frac{\bm{\xi}}{2}\right\}
  \leq&\mathrm{Pr}'\left\{\big|{\bf E}'_N{\bf f}_N-{\bf E}_N{\bf f}_N\big|> \frac{\bm{\xi}}{2} \right\}.
\end{align}
According to Lemma \ref{lem:Chebyshev.A}, since the samples ${\bf z}^{[m]}_n$ ($1\leq m\leq M,\;1\leq n\leq M$) are independent of each other, we have
\begin{align}\label{eq:sym.pr3.A}
&\mathrm{Pr}'\left\{\big|{\bf E}{\bf f}_N-{\bf E}'_N{\bf f}_N\big|\not\leq \frac{\bm{\xi}}{2}\right\}\nonumber\\
=&\mathrm{Pr}\left\{\begin{pmatrix}
                      \Big|\sum_{n=1}^N\big(\mathrm{E}^{[1]}\widehat{f}^{[1]}-\widehat{f}^{[1]}({\bf z}_n^{[1]})\big)\Big| \\
                      \vdots \\
                      \Big|\sum_{n=1}^N\big(\mathrm{E}^{[M]}\widehat{f}^{[M]}-\widehat{f}^{[M]}({\bf z}_n^{[M]})\big)\Big| \\
                    \end{pmatrix}
                    \not\leq
                    \begin{pmatrix}
                      \frac{N\xi^{[1]}}{2} \\
                      \vdots \\
                      \frac{N\xi^{[M]}}{2} \\
                    \end{pmatrix}
\right\}\nonumber\\
\leq&\mathrm{Pr}\left\{\begin{pmatrix}
                      \sum_{n=1}^N\big|\mathrm{E}^{[1]}\widehat{f}^{[1]}-\widehat{f}^{[1]}({\bf z}_n^{[1]})\big| \\
                      \vdots \\
                      \sum_{n=1}^N\big|\mathrm{E}^{[M]}\widehat{f}^{[M]}-\widehat{f}^{[M]}({\bf z}_n^{[M]})\big| \\
                    \end{pmatrix}
                    \not\leq
                    \begin{pmatrix}
                      \frac{N\xi^{[1]}}{2} \\
                      \vdots \\
                      \frac{N\xi^{[M]}}{2} \\
                    \end{pmatrix}
\right\}\nonumber\\
\leq &\sum_{m=1}^M\sum_{\Lambda^{[m]}\subset\bm{\Lambda}}\left(\phi_{\bm{\mathcal{F}}}(\Lambda^{[m]},\bm{\xi})+
  \frac{N\sum\limits_{i\in \Lambda^{[m]}}\mathrm{E}^{[i]}\left\{\big(\mathrm{E}^{[i]}\widehat{f}^{[i]}
  -\widehat{f}^{[i]}({\bf z}^{[i]})\big)^2\right\}}{\frac{N^2}{4}\sum\limits_{i\in \Lambda^{[m]}}(\xi^{[i]})^2}\right)\nonumber\\
  =& \sum_{m=1}^M\sum_{\Lambda^{[m]}\subset\bm{\Lambda}}\left(\phi_{\bm{\mathcal{F}}}(\Lambda^{[m]},\bm{\xi})+
\frac{\sum\limits_{i\in \Lambda^{[m]}}4\mathrm{Var}^{[i]}\big(\widehat{f}^{[i]}({\bf z}^{[i]})\big) }{N\sum\limits_{i\in \Lambda^{[m]}}(\xi^{[i]})^2}\right)\qquad(*)\nonumber\\
  \leq& \sum_{m=1}^M\sum_{\Lambda^{[m]}\subset\bm{\Lambda}}\left(\phi_{\bm{\mathcal{F}}}(\Lambda^{[m]},\bm{\xi})+
\frac{4m(b-a)^2}{N\sum\limits_{i\in \Lambda^{[m]}}(\xi^{[i]})^2}\right),
\end{align}
where the step $(*)$ is followed from the fact that for each task $\mathcal{Z}^{[m]}$ ($1\leq m\leq M$), the samples $\{z_n^{[m]}\}_{n=1}^N$ are independent.

Hence, we get
\begin{align}\label{eq:sym.pr4.A}
 & {\bf 1}_{\left\{|{\bf E}{\bf f}_N-{\bf E}_N{\bf f}_N|> \bm{\xi}\right\}}
 \left(1-\left(\sum_{m=1}^M\sum_{\Lambda^{[m]}\subset\bm{\Lambda}}\phi_{\bm{\mathcal{F}}}(\Lambda^{[m]},\bm{\xi})+
\frac{4m(b-a)^2}{N\sum\limits_{i\in \Lambda^{[m]}}(\xi^{[i]})^2}\right)\right)\nonumber\\
  \leq&\mathrm{Pr}'\left\{\big|{\bf E}'_N{\bf f}_N-{\bf E}_N{\bf f}_N\big|> \frac{\bm{\xi}}{2} \right\}.
\end{align}
Taking the expectation with respect to the sample collection $\{{\bf Z}_N^{[m]}\}_{m=1}^M$ of the tasks $\mathcal{Z}^{[1]},\cdots,\mathcal{Z}^{[M]}$ and letting
 \begin{align}\label{eq:sym.pr5.A}
  \sum_{m=1}^M\sum_{\Lambda^{[m]}\subset\bm{\Lambda}}\bigg(\phi_{\bm{\mathcal{F}}}(\Lambda^{[m]},\bm{\xi})+
\frac{4m(b-a)^2}{N\sum\limits_{i\in \Lambda^{[m]}}(\xi^{[i]})^2}\bigg)\leq \frac{1}{2},
\end{align}
we then have for any $\bm{\xi}>{\bf 0}$,
\begin{equation*}
  \mathrm{Pr}\left\{\sup_{{\bf f} \in\bm{\mathcal{F}}}
\big|{\bf E}{\bf f}-{\bf E}_N{\bf f}\big|> {\bm \xi} \right\}\leq
2  \mathrm{Pr}\left\{\sup_{{\bf f} \in\bm{\mathcal{F}}}
\big|{\bf E}'_N{\bf f}-{\bf E}_N{\bf f}\big|> \frac{{\bm \xi}}{2} \right\}.
\end{equation*}
This completes the proof. \hfill$\blacksquare$

%%%%%

\subsection{Proof of Theorem \ref{thm:sym}}

{\it Proof of Theorem \ref{thm:sym}.} Let ${\bf f}_N=(\widehat{f}_1,\cdots,\widehat{f}_M)^T$ be the vector-valued function achieving the supremum
\begin{equation*}%\label{eq:sym.pr1}
  \sup_{{\bf f} \in\bm{\mathcal{F}}}
\big\{\big|{\bf E}{\bf f}-{\bf E}_N{\bf f}\big|\big\}.
\end{equation*}
Similar to the proof of Theorem \ref{thm:sym.A}, we have
\begin{align}\label{eq:sym.pr2}
  {\bf 1}_{\{|{\bf E}{\bf f}_N-{\bf E}_N{\bf f}_N|> \bm{\xi}\}}
  \mathrm{Pr}'\left\{\big|{\bf E}{\bf f}_N-{\bf E}'_N{\bf f}_N\big|\leq \frac{\bm{\xi}}{2}\right\}
  \leq&\mathrm{Pr}'\left\{\big|{\bf E}'_N{\bf f}_N-{\bf E}_N{\bf f}_N\big|> \frac{\bm{\xi}}{2} \right\}.
\end{align}
According to Lemma \ref{lem:Chebyshev}, we have
\begin{align}\label{eq:sym.pr3}
&\mathrm{Pr}'\left\{\big|{\bf E}{\bf f}_N-{\bf E}'_N{\bf f}_N\big|\not\leq \frac{\bm{\xi}}{2}\right\}\nonumber\\
\leq&\mathrm{Pr}\left\{\begin{pmatrix}
\sum_{n=1}^N\big|\mathrm{E}\widehat{f}_1-\widehat{f}_1({\bf z}_n^{[1]})\big| \\
                      \vdots \\
\sum_{n=1}^N\big|\mathrm{E}\widehat{f}_M-\widehat{f}_M({\bf z}_n^{[M]})\big| \\
                    \end{pmatrix}
                    \not\leq
                    \begin{pmatrix}
                      N\xi^{[1]}/2 \\
                      \vdots \\
                      N\xi^{[M]}/2 \\
                    \end{pmatrix}
\right\}\nonumber\\
\leq &\sum_{m=1}^M\sum_{\Lambda^{[m]}\subset\bm{\Lambda}}\left(\phi_{\bm{\mathcal{F}}}(\Lambda^{[m]},\bm{\xi})+
  \frac{N\sum\limits_{i\in \Lambda^{[m]}}\mathrm{Var}\left(\widehat{f}^{[i]}({\bf z}^{[i]})\right)+2N^2\mathop{\sum\limits_{i_1<i_2}}\limits_{i_1,i_2\in\Lambda^{[m]}}
  \mathrm{Cov}
  \left(\widehat{f}^{(i_1)}({\bf z}^{(i_1)}),\widehat{f}^{(i_2)}({\bf z}^{(i_2)})\right)}{\frac{N^2}{4}\Big(\sum\limits_{i\in \Lambda^{[m]}}\xi^{[i]}\Big)^2}\right)\nonumber\\
  =& \sum_{m=1}^M\sum_{\Lambda^{[m]}\subset\bm{\Lambda}}
  \left(\phi_{\bm{\mathcal{F}}}(\Lambda^{[m]},\bm{\xi})+
\frac{4\sum\limits_{i\in \Lambda^{[m]}}\mathrm{Var}\left(\widehat{f}^{[i]}({\bf z}^{[i]})\right)}{N\Big(\sum\limits_{i\in \Lambda^{[m]}}\xi^{[i]}\Big)^2}
+\frac{8\mathop{\sum\limits_{i_1<i_2}}\limits_{i_1,i_2\in\Lambda^{[m]}}
  \mathrm{Cov}
  \left(\widehat{f}^{(i_1)}({\bf z}^{(i_1)}),\widehat{f}^{(i_2)}({\bf z}^{(i_2)})\right)}{\Big(\sum\limits_{i\in \Lambda^{[m]}}\xi^{[i]}\Big)^2}\right)\nonumber\\
  \leq& \sum_{m=1}^M\sum_{\Lambda^{[m]}\subset\bm{\Lambda}}\left(\phi_{\bm{\mathcal{F}}}
  (\Lambda^{[m]},\bm{\xi})+
\frac{4m(b-a)^2}{N\Big(\sum\limits_{i\in \Lambda^{[m]}}\xi^{[i]}\Big)^2}
+\frac{8\mathop{\sum\limits_{i_1<i_2}}\limits_{i_1,i_2\in\Lambda^{[m]}}
  \mathrm{Cov}_{\bm{\mathcal{F}}}
  \left(i_1,i_2\right)}{\Big(\sum\limits_{i\in \Lambda^{[m]}}\xi^{[i]}\Big)^2}\right)
\end{align}

Moreover, define
\begin{equation*}%\label{eq:Gamma}
  \Gamma_2:=\sum_{m=1}^M\sum_{\Lambda^{[m]}\subset\bm{\Lambda}}
\frac{m(b-a)^2}{\Big(\sum\limits_{i\in \Lambda^{[m]}}\xi^{[i]}\Big)^2},
\end{equation*}
and
\begin{equation*}%\label{eq:Upsilon}
 \Upsilon_2:=\sum_{m=1}^M\sum_{\Lambda^{[m]}\subset\bm{\Lambda}}
 \frac{8\mathop{\sum\limits_{i_1<i_2}}
 \limits_{i_1,i_2\in\Lambda^{[m]}}
  \mathrm{Cov}_{\bm{\mathcal{F}}}
  \left(i_1,i_2\right)}{\Big(\sum\limits_{i\in \Lambda^{[m]}}\xi^{[i]}\Big)^2}.
\end{equation*}
Hence, we get
\begin{align}\label{eq:sym.pr4}
  {\bf 1}_{\{|{\bf E}{\bf f}_N-{\bf E}_N{\bf f}_N|> \bm{\xi}\}}
 \left(1-\left(\frac{4\Gamma_2}{N}+\Upsilon(\bm{\Lambda})+\Upsilon_2\right)\right)
  \leq\mathrm{Pr}'\left\{\big|{\bf E}'_N{\bf f}_N-{\bf E}_N{\bf f}_N\big|> \frac{\bm{\xi}}{2} \right\}.
\end{align}
Taking the expectation with respect to $\{{\bf Z}_N^{[m]}\}_{m=1}^M$ and letting
 \begin{align}\label{eq:sym.pr5}
  \frac{4\Gamma_2}{N}+\Upsilon(\bm{\Lambda})+\Upsilon_2\leq \frac{1}{2},
\end{align}
we then have for any $\bm{\xi}>0$
\begin{equation*}
  \mathrm{Pr}\left\{\sup_{{\bf f} \in\bm{\mathcal{F}}}
\big|{\bf E}{\bf f}-{\bf E}_N{\bf f}\big|> {\bm \xi} \right\}\leq
2  \mathrm{Pr}\left\{\sup_{{\bf f} \in\bm{\mathcal{F}}}
\big|{\bf E}'_N{\bf f}-{\bf E}_N{\bf f}\big|> \frac{{\bm \xi}}{2} \right\}.
\end{equation*}
This completes the proof. \hfill$\blacksquare$

\subsection{Proof of Theorem \ref{thm:bound.A}}

{\it Proof of Theorem \ref{thm:bound.A}.}
For any $1\leq m\leq M$, consider $\{\epsilon^{[m]}_n\}_{n=1}^N$ as independent Rademacher random variables, {\it i.e.}, independent $\{\pm1\}$-valued random variables with equal probability of taking either value.
Given an $\{\epsilon^{[m]}_n\}_{n=1}^{N}$ and a ${\bf Z}^{[m]}_{2N}$, denote
\begin{align}\label{eq:epsilon}
    \overrightarrow{\epsilon}^{[m]}:=&(\epsilon^{[m]}_1,\cdots,\epsilon^{[m]}_{N},
    -\epsilon^{[m]}_1,\cdots,-\epsilon^{[m]}_{N})^T
\in\{\pm 1\}^{2N},\quad 1\leq m\leq M,
\end{align}
and for any ${\bf f}=(f_1,\cdots,f_M)^T\in\bm{\mathcal{F}}_c^{\mathrm{R}}$,
\begin{align}\label{eq:vecf}
  \overrightarrow{f}^{[m]}({\bf Z}^{[m]}_{2N}):=&\big(f^{[m]}({\bf z'}^{[m]}_1),\cdots,f^{[m]}({\bf z'}^{[m]}_{N}),f^{[m]}({\bf z}^{[m]}_1),\cdots,f^{[m]}({\bf z}^{[m]}_{N})\big)^T\in[a,b]^{2N}.
\end{align}

According to Theorem \ref{thm:sym.A}, given any $\bm{\xi}>{\bf 0}$ and for any $N\in\mathbb{N}$ satisfying Condition \eqref{eq:sym.condition.A},
we have
\begin{align}\label{eq:bas1.A}
    &\mathrm{Pr}\left\{\sup_{{\bf f} \in\bm{\mathcal{F}}_c^{\mathrm{R}}}
\big|{\bf E}{\bf f}-{\bf E}_N{\bf f}\big|> {\bm \xi} \right\}\nonumber\\
\leq& 2\mathrm{Pr}\left\{\sup_{{\bf f} \in\bm{\mathcal{F}}_c^{\mathrm{R}}}
\big|{\bf E}'_N{\bf f}-{\bf E}_N{\bf f}\big|> \frac{{\bm \xi}}{2} \right\}\qquad\mbox{(by Theorem \ref{thm:sym})}\nonumber\\
=&2\mathrm{Pr}\left\{\sup_{{\bf f} \in\bm{\mathcal{F}}_c^{\mathrm{R}}}
\begin{pmatrix}
\Big|\frac{1}{N}\sum_{n=1}^{N}\big(f({\bf z'}^{[1]}_n)-f({\bf z}^{[1]}_n)\big)\Big| \\
\vdots \\
\Big|\frac{1}{N}\sum_{n=1}^{N}\big(f({\bf z'}^{[M]}_n)-f({\bf z}^{[M]}_n)\big)\Big| \\
\end{pmatrix}
>\begin{pmatrix}
  \frac{\xi^{[1]}}{2} \\
  \vdots \\
   \frac{\xi^{[M]}}{2}\\
 \end{pmatrix}
\right\}\nonumber\\
=&2\mathrm{Pr}\left\{\sup_{{\bf f} \in\bm{\mathcal{F}}_c^{\mathrm{R}}}
\begin{pmatrix}
\Big|\frac{1}{N}\sum_{n=1}^{N}\epsilon_n^{[1]}\big(f({\bf z'}^{[1]}_n)-f({\bf z}^{[1]}_n)\big)\Big| \\
\vdots \\
\Big|\frac{1}{N}\sum_{n=1}^{N}\epsilon_n^{[M]}\big(f({\bf z'}^{[M]}_n)-f({\bf z}^{[M]}_n)\big)\Big| \\
\end{pmatrix}
>\begin{pmatrix}
  \frac{\xi^{[1]}}{2} \\
  \vdots \\
   \frac{\xi^{[M]}}{2}\\
 \end{pmatrix}
\right\}\nonumber\\
=&2\mathrm{Pr}\left\{\sup_{{\bf f} \in\bm{\mathcal{F}}^{\mathrm{R}}_c}
\begin{pmatrix}
\Big|\frac{1}{2N}\big\langle\overrightarrow{\epsilon}^{[1]},
\overrightarrow{f}_1({\bf Z}^{[1]}_{2N})\big\rangle\Big| \\
\vdots \\
\Big|\frac{1}{2N}\big\langle\overrightarrow{\epsilon}^{[M]},
\overrightarrow{f}_M({\bf Z}^{[M]}_{2N})\big\rangle\Big| \\
\end{pmatrix}
>\begin{pmatrix}
  \frac{\xi^{[1]}}{4} \\
  \vdots \\
   \frac{\xi^{[M]}}{4}\\
 \end{pmatrix}
\right\}.
\end{align}

For any given sample collection $\{{\bf Z}_{2N}^{[m]}\}_{m=1}^M$ of the tasks $\mathcal{Z}^{[1]},\cdots,\mathcal{Z}^{[M]}$, let $\bm{\Omega}_{p,N}(\bm{\mathcal{F}}^{\mathrm{R}}_c,\bm{\xi}/8)$ be the cover of $\bm{\mathcal{F}}^{\mathrm{R}}_c$ w.r.t. the radius-vectors $\bm{\xi}/8$. Since $\bm{\mathcal{F}}^{\mathrm{R}}_c$ is composed of the functions with the range $[a,b]$, we assume that the same holds for any ${\bf h}\in\bm{\Omega}_{p,N}(\bm{\mathcal{F}}^{\mathrm{R}}_c,\bm{\xi}/8)$.
If ${\bf f}_{\dag}=(f_{\dag}^{[1]},\cdots,f_{\dag}^{[M]})^T$ is a vector-valued function that achieves
\begin{equation*}
 \sup_{{\bf f} \in\bm{\mathcal{F}}^{\mathrm{R}}_c}
\begin{pmatrix}
\Big|\frac{1}{2N}\big\langle\overrightarrow{\epsilon}^{[1]},
\overrightarrow{f}^{[1]}({\bf Z}^{[1]}_{2N})\big\rangle\Big| \\
\vdots \\
\Big|\frac{1}{2N}\big\langle\overrightarrow{\epsilon}^{[M]},
\overrightarrow{f}^{[M]}({\bf Z}^{[M]}_{2N})\big\rangle\Big| \\
\end{pmatrix}
>\begin{pmatrix}
  \frac{\xi^{[1]}}{4} \\
  \vdots \\
   \frac{\xi^{[M]}}{4}\\
 \end{pmatrix},
\end{equation*}
 there must be an ${\bf h}_{\dag}=(h_{\dag}^{[1]},\cdots,h_{\dag}^{[M]})^T\in\bm{\Omega}_{p,N}
 (\bm{\mathcal{F}}^{\mathrm{R}}_c,\bm{\xi}/8)$ such that, for any $1\leq m\leq M$,
\begin{align*}
    \frac{1}{2N}\sum_{n=1}^{N}\left(|f^{[m]}_{\dag}({\bf z'}^{[m]}_n)-h^{[m]}_{\dag}({\bf z'}^{[m]}_n)|+|f^{[m]}_{\dag}({\bf z}^{[m]}_n)-h^{[m]}_{\dag}({\bf z}^{[m]}_n)|\right)< \frac{\xi^{[m]}}{8},
\end{align*}
and meanwhile,
\begin{equation*}
    \Big|\frac{1}{2N}\big\langle\overrightarrow{\epsilon}^{[m]},
\overrightarrow{h}_{\dag}^{[M]}({\bf Z}^{[m]}_{2N})\big\rangle\Big|>\frac{\xi^{[m]}}{8}.
\end{equation*}
Therefore, we arrive at
\begin{align}\label{eq:bas2.A}
&\mathrm{Pr}\left\{\sup_{{\bf f} \in  \bm{\mathcal{F}}^{\mathrm{R}}_c }
\begin{pmatrix}
\Big|\frac{1}{2N}\big\langle\overrightarrow{\epsilon}^{[1]},
\overrightarrow{f}^{[1]}({\bf Z}^{[1]}_{2N})\big\rangle\Big| \\
\vdots \\
\Big|\frac{1}{2N}\big\langle\overrightarrow{\epsilon}^{[M]},
\overrightarrow{f}^{[M]}({\bf Z}^{[M]}_{2N})\big\rangle\Big| \\
\end{pmatrix}
>\begin{pmatrix}
  \frac{\xi^{[1]}}{4} \\
  \vdots \\
   \frac{\xi^{[M]}}{4}\\
 \end{pmatrix}
\right\}\nonumber\\
\leq&\mathrm{Pr}\left\{\sup_{{\bf h} \in\bm{\Omega}_{p,N}(\bm{\mathcal{F}}^{\mathrm{R}}_c,\bm{\xi}/8)}
\begin{pmatrix}
\Big|\frac{1}{2N}\big\langle\overrightarrow{\epsilon}^{[1]},
\overrightarrow{h}^{[1]}({\bf Z}^{[1]}_{2N})\big\rangle\Big| \\
\vdots \\
\Big|\frac{1}{2N}\big\langle\overrightarrow{\epsilon}^{[M]},
\overrightarrow{h}^{[M]}({\bf Z}^{[M]}_{2N})\big\rangle\Big| \\
\end{pmatrix}
>\begin{pmatrix}
  \frac{\xi^{[1]}}{8} \\
  \vdots \\
   \frac{\xi^{[M]}}{8}\\
 \end{pmatrix}
\right\}.
\end{align}

On the other hand, given a $\bm{\xi}>{\bf 0}$ and for any $N\in\mathbb{N}$ satisfying Condition \eqref{eq:sym.condition.A},
\begin{align}\label{eq:bas3.A}
&\mathrm{Pr}\left\{\sup_{{\bf h} \in\bm{\Omega}_{p,N}(\bm{\mathcal{F}}^{\mathrm{R}}_c,\bm{\xi}/8)}
\begin{pmatrix}
\Big|\frac{1}{2N}\big\langle\overrightarrow{\epsilon}^{[1]},
\overrightarrow{h}^{[1]}({\bf Z}^{[1]}_{2N})\big\rangle\Big| \\
\vdots \\
\Big|\frac{1}{2N}\big\langle\overrightarrow{\epsilon}^{[M]},
\overrightarrow{h}^{[M]}({\bf Z}^{[M]}_{2N})\big\rangle\Big| \\
\end{pmatrix}
>\begin{pmatrix}
  \frac{\xi^{[1]}}{8} \\
  \vdots \\
   \frac{\xi^{[M]}}{8}\\
 \end{pmatrix}
\right\}\nonumber\\
=&\mathrm{Pr}\left\{\sup_{{\bf h} \in\bm{\Omega}_{p,N}(\bm{\mathcal{F}}^{\mathrm{R}}_c,\bm{\xi}/8)}
\begin{pmatrix}
\Big|\frac{1}{N}\big\langle\overrightarrow{\epsilon}^{[1]},
\overrightarrow{h}^{[1]}({\bf Z}^{[1]}_{2N})\big\rangle\Big| \\
\vdots \\
\Big|\frac{1}{N}\big\langle\overrightarrow{\epsilon}^{[M]},
\overrightarrow{h}^{[M]}({\bf Z}^{[M]}_{2N})\big\rangle\Big| \\
\end{pmatrix}
>\begin{pmatrix}
  \frac{\xi^{[1]}}{4} \\
  \vdots \\
   \frac{\xi^{[M]}}{4}\\
 \end{pmatrix}
\right\}\nonumber\\
=&\mathrm{Pr}\left\{\sup_{{\bf h}\in\bm{\Omega}_{p,N}(\bm{\mathcal{F}}^{\mathrm{R}}_c,\bm{\xi}/8)}\big|{\bf E}'_N{\bf h}
-{\bf E}_N{\bf h}\big|>\frac{\bm{\xi}}{4}\right\}\quad \mbox{(similer to \eqref{eq:bas1.A})}\nonumber\\
\leq&\mathrm{Pr}\left\{\sum_{{\bf h}\in\bm{\Omega}_{p,N}(\bm{\mathcal{F}}^{\mathrm{R}}_c,\bm{\xi}/8)}\big|{\bf E}'_N{\bf h}
-{\bf E}_N{\bf h}\big|>\frac{\bm{\xi}}{4}\right\}\nonumber\\
\leq&\mathrm{Pr}\left\{\sum_{{\bf h}\in\bm{\Omega}_{p,N}(\bm{\mathcal{F}}^{\mathrm{R}}_c,\bm{\xi}/8)}\big|{\bf E}{\bf h}
-{\bf E}_N{\bf h}\big|+\big|{\bf E}{\bf h}
-{\bf E}'_N{\bf h}\big|>\frac{\bm{\xi}}{4}\right\}\nonumber\\
\leq&2\mathrm{Pr}\left\{\sum_{{\bf h}\in\bm{\Omega}_{p,N}(\bm{\mathcal{F}}^{\mathrm{R}}_c,\bm{\xi}/8)}\big|{\bf E}{\bf h}
-{\bf E}_N{\bf h}\big|
>\frac{\bm{\xi}}{8}\right\}\nonumber\\
\leq&2^{M+1} \bm{\mathcal{N}}_1\big(\bm{\mathcal{F}}_c^{\mathrm{R}},\bm{\xi}/8,2N\big) \exp\left\{\frac{-N\sum_{m=1}^M(\xi^{[m]})^2}{32M^2(b-a)^2}\right\}.
\end{align}
The last inequality of \eqref{eq:bas3.A} is derived from Definition \eqref{def:UEN} and Theorem \ref{thm:dev}.

%
%Moreover, we denote the event
%\begin{equation*}%\label{eq:event}
%A:=\left\{\sup_{h\in\Lambda}\Big|\frac{1}{2N}
%\big\langle\overrightarrow{\epsilon},
%\overrightarrow{h}({\bf Z}_{1}^{2N})\big\rangle\Big|>\frac{\xi}{8}\right\},
%\end{equation*}
%and let $\chi_A$ be the characteristic function of the event $A$. By Fubini's Theorem, we have
%\begin{align}\label{eq:Fubini}
%&\mathrm{Pr}\{A\}=\mathrm{E}_{{\bf Z}_1^{2N}}\Big\{\mathrm{E}_{\overrightarrow{\epsilon}}
%\big\{\chi_A\big\}\Big\}\\
%=&\mathrm{E}_{{\bf Z}_1^{2N}}\left\{\mathrm{Pr}\left\{\sup_{h\in\Lambda}\Big|\frac{1}{2N}\big\langle\overrightarrow{\epsilon},
%\overrightarrow{h}({\bf Z}_{1}^{2N})\big\rangle\Big|>\frac{\xi}{8}\right\}\right\}.\nonumber
%\end{align}
%

The combination of \eqref{eq:bas1.A}, \eqref{eq:bas2.A} and \eqref{eq:bas3.A} leads to the result: given any $\bm{\xi}>{\bf 0}$, there holds that for any $N\in\mathbb{N}$ satisfying Condition \eqref{eq:sym.condition.A},
\begin{align*}%\label{eq:RB1_1}
     \mathrm{Pr}\left\{\sup_{{\bf f} \in\bm{\mathcal{F}}_c^\mathrm{R}}
\big|{\bf E}{\bf f}-{\bf E}_N{\bf f}\big|> {\bm \xi} \right\}\leq 2^{M+2} \bm{\mathcal{N}}_1\big(\bm{\mathcal{F}}_c^{\mathrm{R}},\bm{\xi}/8,2N\big) \exp\left\{\frac{-N\sum_{m=1}^M(\xi^{[m]})^2}{32M^2(b-a)^2}\right\}.
\end{align*}
This completes the proof. \hfill$\blacksquare$

\subsection{Proof of Theorem \ref{thm:small.prob}}

Before the formal proof, we present a necessary lemma.

\begin{lemma}\label{lem:Con.Markov}
Let ${\bf s}_n=(s^{[1]}_n,\cdots,s^{[M]}_n)\in\mathbb{R}^M$ ($1\leq n\leq N$) be $N$ i.i.d. random vectors. Then, there holds that for any $\bm{\xi}=(\xi^{[1]},\cdots,\xi^{[M]})^T>{\bf 0}$,
\begin{align}\label{eq:Con.Markov}
  \mathrm{Pr}\left\{ \sum_{n=1}^N {\bf s}_n  \leq N{\bm \xi} \right\}\leq
2^M\mathrm{Pr}\Big\{
{\bf s}_1\leq 2{\bm \xi} \Big\}.
\end{align}
\end{lemma}

{\it Proof.} For any $1\leq m\leq M$, we have
\begin{equation*}
\sum_{n=1}^Ns_n^{[m]}\geq \sum_{n=1}^Ns_n^{[m]}{\bf 1}_{\left\{s_n^{[m]}> 2\xi^{[m]}\right\}}
\geq 2\xi^{[m]}\sum_{n=1}^N{\bf 1}_{\left\{s_n^{[m]}> 2\xi^{[m]}\right\}}.
\end{equation*}
Hence, it is followed from the conditional Markov inequality that
\begin{align*}
\mathrm{Pr}\left\{ \sum_{n=1}^N {\bf s}_n  \leq N{\bm \xi} \right\}
\leq&\mathrm{Pr}\left\{
                               \begin{pmatrix}
                                 2\xi^{[1]}\sum_{n=1}^N{\bf 1}_{\left\{s_n^{[1]}> 2\xi^{[1]}\right\}}\\
                                  \vdots \\
                                   2\xi^{[M]}\sum_{n=1}^N{\bf 1}_{\left\{s_n^{[M]}> 2\xi^{[M]}\right\}}\\
                                \end{pmatrix}\leq  N \begin{pmatrix}
                                 \xi^{[1]}\\
                                  \vdots \\
                                   \xi^{[M]}\\
                                \end{pmatrix}\right\} \nonumber\\
    =&    \mathrm{Pr}\left\{
                               \begin{pmatrix}
                                 \sum_{n=1}^N{\bf 1}_{\left\{s_n^{[1]}\leq 2\xi^{[1]}\right\}}\\
                                  \vdots \\
                                   \sum_{n=1}^N{\bf 1}_{\left\{s_n^{[M]}\leq 2\xi^{[M]}\right\}}\\
                                \end{pmatrix}\geq  (1-2^{-1})N \begin{pmatrix}
                                 1\\
                                  \vdots \\
                                  1\\
                                \end{pmatrix}\right\} \nonumber\\
      =&  \mathrm{Pr}\left\{ \sum_{n=1}^N{\bf 1}_{\left\{s_n^{[1]}\leq 2\xi^{[1]}\right\}} \geq 2^{-1}N\;\big| \;\mathcal{A}_2^M  \right\}    \mathrm{Pr}\left\{ \mathcal{A}_2^M  \right\} \nonumber\\
 \leq & \frac{\mathrm{E}\left\{ \sum_{n=1}^N{\bf 1}_{\left\{s_n^{[1]}\leq 2\xi^{[1]}\right\}} \;\big| \;\mathcal{A}_2^M  \right\}}{2^{-1}N}  \mathrm{Pr}\left\{ \mathcal{A}_2^M  \right\} \nonumber\\
 \leq & \frac{N\mathrm{Pr}\left\{ s_1^{[1]}\leq 2\xi^{[1]} \;\big| \;\mathcal{A}_2^M  \right\}}{2^{-1}N}  \mathrm{Pr}\left\{ \mathcal{A}_2^M  \right\}
 = 2\mathrm{Pr}\left\{ s_1^{[1]}\leq 2\xi^{[1]} ,\mathcal{A}_2^M  \right\},
\end{align*}
where $\mathcal{A}_2^M$ stands for the event that $\Big\{ \sum_{n=1}^N{\bf 1}_{\{s_n^{[m]}\leq 2\xi^{[m]}\}}  \Big\}_{m=2}^M$. Then, following this way, we have
\begin{align*}
&\mathrm{Pr}\left\{ \sum_{n=1}^N {\bf s}_n  \leq N{\bm \xi} \right\}
\leq 2\mathrm{Pr}\left\{ s_1^{[1]}\leq 2\xi^{[1]} ,\mathcal{A}_2^M  \right\}\leq 2^2
\mathrm{Pr}\left\{ s_1^{[1]}\leq 2\xi^{[1]} ,s_1^{[2]}\leq 2\xi^{[2]},\mathcal{A}_3^M  \right\}
\nonumber\\
 &\leq \cdots\leq 2^M\mathrm{Pr}\left\{ s_1^{[1]}\leq 2\xi^{[1]} ,s_1^{[2]}\leq 2\xi^{[2]},
\cdots, s_1^{[M]}\leq 2\xi^{[M]}  \right\}=2^M\mathrm{Pr}\left\{ {\bf s}_1\leq 2\bm{\xi}  \right\}.
\end{align*}
This completes the proof. \hfill$\blacksquare$

Next, we come up with the proof of Theorem \ref{thm:small.prob}.

{\it Proof of Theorem \ref{thm:small.prob}.} Let $\widehat{{\bf f}}_*=(f_*^{[1]},\cdots,f_*^{[M]})^T$
be the vector-valued function achieving the supremum $\sup_{{\bf f} \in\bm{\mathcal{F}}_c^\mathrm{R}}
\big|{\bf E}{\bf f}-{\bf E}_N{\bf f}\big|$. Then, it is followed from Lemma \ref{lem:Con.Markov} that
\begin{align*}
\mathrm{Pr}\left\{\sup_{{\bf f} \in\bm{\mathcal{F}}_c^\mathrm{R}}
\big|{\bf E}{\bf f}-{\bf E}_N{\bf f}\big|\leq {\bm \xi} \right\}
=&\mathrm{Pr}\left\{\big|{\bf E}{\bf f}_*-{\bf E}_N{\bf f}_*\big|\leq {\bm \xi} \right\}\leq2^M\mathrm{Pr}\left\{{\bf s}_*\leq 2{\bm \xi} \right\}\nonumber\\
\leq &2^M\sup_{{\bf f} \in\bm{\mathcal{F}}_c^\mathrm{R}}\mathrm{Pr}\big\{
{\bf s}\leq 2{\bm \xi} \big\},
\end{align*}
where ${\bf s}_*=\big(s_*^{[1]},\cdots,s_*^{[M]}\big)^T$ with $s_*^{[m]}:=\big|\mathrm{E}^{[m]}f_*^{[m]}
-f_*^{[m]}({\bf z}^{[m]})\big|$ for any $1\leq m\leq M$.
\hfill$\blacksquare$

\subsection{Proof of Theorem \ref{thm:main}}

{\it Proof of Theorem \ref{thm:main}.} Denote ${\bf t}_N=(t^{[1]},\cdots,t^{[M]})^T$ with $t^{[i]}:=\big|\mathrm{E}^{[i]}f^{[i]}
-\mathrm{E}_N^{[i]}f^{[i]}\big|>\xi^{[i]}$. The event ${\bf t}_N\not\leq {\bm \xi}$ contains the following possibilities:
\begin{itemize}
  \item $\mathcal{P}^{[1]}$: there is only one index $\{i\}= \Lambda^{[1]}\subset\bm{\Lambda}$ satisfying that $t^{[i]}>\xi^{[i]}$;
  \item $\mathcal{P}^{[m]}$: there are $m$ ($1<m<M$) indices $\{i^{[1]},\cdots,i^{[m]}\}=\Lambda^{[m]}\subset\bm{\Lambda}$ satisfying that $t^{[i_k]}>\xi^{[i_k]}$ ($1\leq k\leq m$);
  \item $\mathcal{P}^{[M]}$: $t^{[m]}>\xi^{[m]}$ holds for any $1\leq m\leq M$.
\end{itemize}
Thus, we have
\begin{equation}\label{eq:main.pr1}
  \mathrm{Pr}\left\{   {\bf t}_N\not\leq \bm{\xi} \right\}= \mathrm{Pr}\{\mathcal{P}^{[1]}\}+\cdots+\mathrm{Pr}\{\mathcal{P}^{[M]}\}.
\end{equation}
Then, the combination of Definition \ref{def:EDDM}, Theorems \ref{thm:bound.A}\&\ref{thm:small.prob} and \eqref{eq:main.pr1} leads to the result \eqref{eq:main}. Moreover, since $\mathrm{Pr}\left\{ \big\{ s^{[\lambda]}\leq2\xi^{[\lambda]} \big\}_{\lambda\in\overline{\Lambda^{[m]}}}  \right\}\leq 1$ holds for any $\Lambda^{[m]}\subset\bm{\Lambda}$, the result \eqref{eq:main2} can be directly obtained. This completes the proof. \hfill$\blacksquare$

%--------------------------------------------------

%\section*{Acknowledgment}

%-----------------------------------------------

%\bibliographystyle{IEEEtran}

\end{document}